%% file: kvfocus_neurips2026_camera_ready_v2.tex
\documentclass{article}

\PassOptionsToPackage{numbers,compress}{natbib}

\usepackage[main,final]{neurips_2026}

\usepackage[utf8]{inputenc}
\usepackage[T1]{fontenc}
\usepackage{url}
\usepackage{booktabs}
\usepackage{amsfonts,amsmath,amssymb,amsthm}
\usepackage{nicefrac}
\usepackage{microtype}
\usepackage{xcolor}
\usepackage{graphicx}
\usepackage{float}
\usepackage{mathtools}
\usepackage{algorithm}
\usepackage{algpseudocode}
\usepackage{bm}
\usepackage{multirow}
\usepackage{enumitem}
\usepackage{array}
\usepackage{makecell}
\usepackage{xspace}
\usepackage{subcaption}

\usepackage[breaklinks=true]{hyperref}

\newcommand{\lcheck}{\ell^{\star}}

\newcommand{\margint}{m_t}

\newcommand{\Pseg}{P}
\newcommand{\Cseg}{C}
\newcommand{\Sseg}{S}
\newcommand{\Spsuff}{S_p^{\text{suff}}}

\newtheorem{proposition}{Proposition}

\newtheorem{assumption}{Assumption}

\theoremstyle{remark}

\title{KVFocus: A Perturbation-Theoretic Token-Risk Score for Selective KV Cache Reuse in RAG}
\author{%
  \textbf{Shizhuo Zhang}$^{1}$ \quad
  \textbf{Nuowen Kan}$^{2}$\thanks{Correspondence to: Nuowen Kan \textless nwkan@ee.ecnu.edu.cn\textgreater.} \quad
  \textbf{Chenglin Li}$^{1}$ \quad
  \textbf{Ruixiao Zhang}$^{3}$ \quad
  \textbf{Yifan Zhao}$^{1}$ \\
  \textbf{Yuanpeng He}$^{4}$ \quad
  \textbf{Huixin Zhang}$^{5}$ \quad
  \textbf{Fan He}$^{5}$ \quad
  \textbf{Hang Xu}$^{5}$ \quad
  \textbf{Hao Zhang}$^{2}$ \\
  \textbf{Wenrui Dai}$^{1}$ \quad
  \textbf{Junni Zou}$^{1}$ \quad
  \textbf{Hongkai Xiong}$^{1}$ \\
  $^{1}$Shanghai Jiao Tong University \quad
  $^{2}$East China Normal University \\
  $^{3}$University of Illinois Urbana-Champaign, \quad
  $^{4}$School of Computer Science, Peking University \\
  $^{5}$Yinwang Intelligent Technology Co. Ltd. \\
}

\begin{document}

\maketitle

\begin{abstract}
In retrieval-augmented generation (RAG) scenarios, the prefill delay of long inputs to large language models (LLMs) is the dominant component of the time-to-first-token (TTFT). Reusing KV caches across different LLM inputs reduces this cost, but the resulting cross-context mismatch on reused chunks significantly degrades answer quality as it propagates through prefill attention into decode-time logits. Existing selective-recomputation methods mitigate this issue by recomputing tokens chosen by empirical importance heuristics, which overlook the propagation of cache mismatches into answer-side errors, thereby failing to effectively strike the quality-TTFT trade-off. To address this issue, we propose a perturbation-theoretic KV cache reuse framework, \textit{KVFocus}, which selects the reused tokens to recompute through a token-risk score during the prefill process. Specifically, we first develop a first-order perturbation analysis that traces reuse error in three stages: source mismatch on the reused region, suffix drift during prefill, and decode-time logit stability. This analysis reveals that the contribution of each reused token to the suffix-side drift is the product of its source-side cache deviation and the attention mass it receives from the suffix. Guided by this multiplicative structure, we define a token-risk score that can be computed at a single check layer. A plug-and-play selector is designed for existing RAG-based LLM serving systems to select the top-$r\%$ tokens ranked by this score for recomputation in subsequent layers of the same prefill pass. Empirical results across three 7B or 8B instruction-tuned LLMs (Mistral, Qwen2.5, Qwen3) and seven datasets demonstrate that KVFocus achieves a favorable trade-off between answer quality and TTFT compared with state-of-the-art baselines. Compared to full recomputation, KVFocus achieves an average $2.22\times$ TTFT speedup at the cost of an $18\%$ quality degradation.
\end{abstract}

% ============================================================
% ABSTRACT
% ============================================================

\makeatletter
\let\kvfocusOriginalSection\section
\let\kvfocusOriginalSubsection\subsection
\renewcommand{\section}{%
  \@startsection{section}{1}{\z@}%
    {-1.7ex \@plus -0.5ex \@minus -0.2ex}%
    {1.3ex \@plus 0.3ex \@minus 0.2ex}%
    {\large\bf\raggedright}%
}
\renewcommand{\subsection}{%
  \@startsection{subsection}{2}{\z@}%
    {-1.5ex \@plus -0.5ex \@minus -0.2ex}%
    {0.6ex \@plus 0.2ex}%
    {\normalsize\bf\raggedright}%
}
\makeatother

\section{Introduction}
\label{sec:intro}

Retrieval-augmented generation (RAG) addresses the knowledge limitation of large language models (LLMs) by retrieving relevant documents and prepending them to the input prompt~\citep{lewis2020retrieval,gao2023retrieval}. For typical RAG deployments, the prompt length is dominated by retrieved documents, making the prefill stage the dominant cost of time-to-first-token (TTFT)~\citep{kwon2023efficient,yao2025cacheblend}. Cross-request \emph{KV cache reuse}~\citep{gim2024prompt,zheng2024sglang} amortizes this cost by reusing the key (K) and value (V) tensors of cache-hit document chunks computed offline rather than recomputing them online.
However, the naive KV cache reuse strategy suffers from \emph{cached-chunk mismatch}~\citep{yao2025cacheblend}: the cached K/V of a chunk was computed under one offline encoding context, while the chunk now appears under a different online prefix, so the reused KV misses the cross-attention with the preceding online context. This mismatch propagates through prefill attention into decode-time logits, potentially degrading the final answer quality.

\begin{figure}[!t]
  \centering
  \begin{subfigure}[b]{0.40\textwidth}
    \centering
    \includegraphics[width=\linewidth]{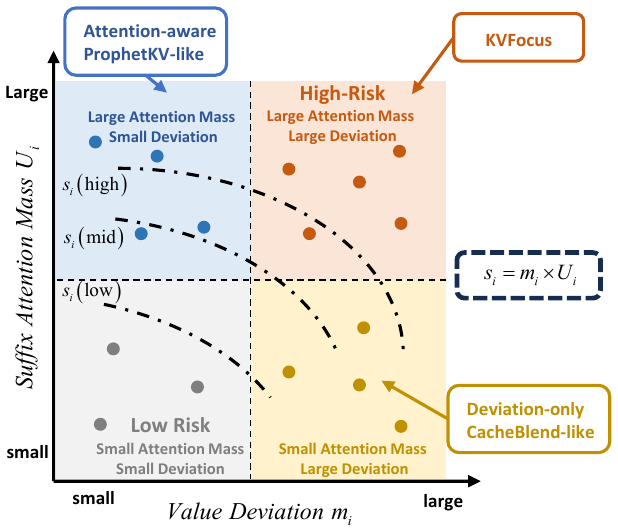}
    \caption{Two-factor risk structure.}
    \label{fig:teaser_a}
  \end{subfigure}%
  \hspace{0.04\textwidth}%
  \begin{subfigure}[b]{0.35\textwidth}
    \centering
    \includegraphics[width=\linewidth]{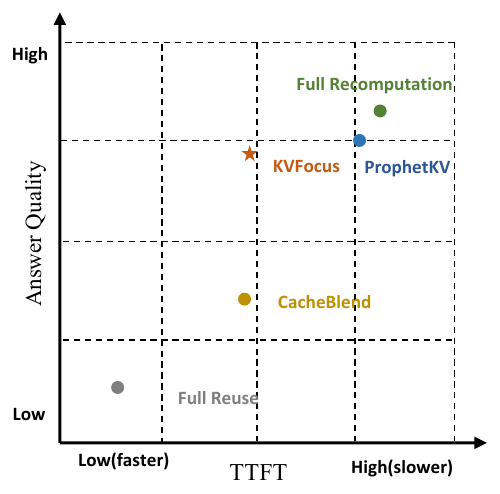}
    \caption{Quality-TTFT trade-off.}
    \label{fig:teaser_b}
  \end{subfigure}
  \caption{
  (\subref{fig:teaser_a}) Only tokens that are both \emph{deviated} (large $m_i$) and \emph{heavily attended} by the suffix (large $U_i$) substantially perturb the answer. Deviation-only methods (CacheBlend-like) rank tokens along the x-axis only; attention-aware methods (ProphetKV-like) rank along the y-axis only; KVFocus ranks by the product $s_i = m_i \cdot U_i$, with the dashed curves marking equal-score level sets. (\subref{fig:teaser_b}) Qualitative illustration of the quality-TTFT trade-off observed in our experiments (Section~\ref{sec:exp_matched_rate}).}\vspace{-0.4cm}
  \label{fig:teaser}
\end{figure}

Recent works mitigate this issue by \emph{selective recomputation}: they choose a small fraction of cached tokens to recompute online while keeping the rest reused. These studies, including CacheBlend~\citep{yao2025cacheblend}, EPIC~\citep{hu2024epic}, and ProphetKV~\citep{wang2026prophetkv}, rank tokens by intermediate signals, such as shallow-layer KV deviations, document boundaries, query relevance, or attention statistics. A detailed discussion of related work is given in Appendix~\ref{app:related_landscape}. Although each achieves gains over the naive full reuse method, all of them rely on a single signal to decide which tokens are worth recomputing. This is insufficient because a reused token harms the final answer only when \emph{both} of the following conditions hold: \textit{i)~the token is deviated:}~its cached K/V differs from what online recomputation would produce, because the KV cache was built without seeing the current online prefix; \textit{ii)~the token is heavily attended:}~downstream suffix tokens assign it a large attention mass, so the deviation propagates into the hidden states used to start generation. A deviated token that the suffix barely attends to consumes recomputation budget without improving answer quality. In contrast, a heavily attended token whose cache is accurate does not need to be recomputed. Existing methods select the tokens to recompute based on at most one of these two conditions (Fig.~\ref{fig:teaser}(\subref{fig:teaser_a})), so part of their recomputation budget is spent on tokens that violate the other condition, which degrades the quality-TTFT trade-off.

To strike an optimal quality-TTFT trade-off, we propose \textit{\textbf{KVFocus}}, which selects the reused tokens to recompute with a token-risk score that estimates how much each reused token harms the final answer. Specifically, we trace the answer-side error caused by a reused token through three stages: \textit{i)} the cache mismatch on the reused region, \textit{ii)} its propagation through prefill attention to the downstream suffix, and \textit{iii)} the resulting logit drift during decoding. Under a first-order approximation, the contribution of a reused token to the suffix-side drift is the product of a source-side factor (how much the cached value deviates) and a downstream factor (how strongly the suffix attends to it), yielding a per-token risk score. KVFocus computes the score for all reused tokens at a single check layer and recomputes the top-$r\%$ of them in the subsequent layers. Our contributions are summarized as follows:

\begin{itemize}[leftmargin=*,itemsep=2pt,topsep=2pt]
  \item We trace how a reused token's cache mismatch propagates through prefill attention into decode-time logits, and show that its contribution to the suffix-side drift is the product of a source-side cache-deviation factor and a downstream attention-mass factor. To our knowledge, this is the first quantitative connection between cache-time mismatch and answer-side error in non-prefix KV cache reuse. Rather than providing a tight bound, the analysis identifies which measurable quantities govern the reuse error and how they combine, which explains why single-factor heuristics can waste recomputation budget.
  \item We propose a per-token risk score that adopts the multiplicative error structure to simultaneously capture how much the reused KV cache deviates and how strongly the suffix attends to the reused token. We then design a training-free, plug-and-play selector for existing RAG serving systems. The selector computes both factors at a single check layer and recomputes the top-ranked tokens in later layers, without requiring an additional forward pass.
  \item We implement KVFocus with the SGLang serving framework~\citep{zheng2024sglang}, and evaluate it across three 7B or 8B instruction-tuned LLMs (Mistral~\citep{jiang2023mistral7b}, Qwen2.5~\citep{qwen2025qwen25technicalreport}, and Qwen3~\citep{yang2025qwen3}) and seven public datasets that include three task types. KVFocus achieves a better quality-TTFT trade-off than existing selective-recomputation baselines. Compared to full recomputation, it achieves an average $2.22\times$ TTFT speedup at the cost of an $18\%$ quality degradation. Our implementation is available at
\url{https://github.com/zhuo121/KVFocus}.
\end{itemize}

\section{Theoretical Framework: From Cached-Chunk Mismatch to Logit Drift}
\label{sec:theory}

We develop a first-order perturbation analysis tracing how reuse error propagates from cached-chunk mismatch on a reused region, through suffix drift during prefill, to decode-time logit drift.
Perturbative and gradient-based analyses have been widely used to study attention behavior in transformers~\citep{abnar2020quantifying,kobayashi2020attention}, and informal first-order arguments underlie recent selective-recomputation methods~\citep{yao2025cacheblend,hu2024epic}.
To our knowledge, this is the first explicit first-order treatment of cross-context KV cache reuse error in RAG that traces mismatches from cached chunks through suffix drift to decode-time logits.
The bounds in this section are first-order in the cache mismatch: higher-order terms are collected in remainder terms. We therefore use these results as design motivation rather than as tight guarantees: they identify which measurable quantities enter the first-order value term, namely value deviation and attention mass, and show that they enter as a product rather than a sum.

% --- 3.1 Setup ---
\subsection{Problem Formulation}
\label{sec:theory:setup}

For a RAG request that concatenates one or more retrieved chunks, we decompose the prompt $X$ at the token level as
\begin{equation}
\label{eq:decomp}
X \;=\; \Pseg \,\sqcup\, \Cseg \,\sqcup\, \Sseg,
\end{equation}
where $\Pseg$ is the online-computed prefix containing at least the system prompt, $\Cseg$ is the set of reused token positions, which may span one or more cached chunks, and $\Sseg$ is the downstream online suffix containing the question. The symbol $\sqcup$ denotes concatenation of disjoint token-position sets in prompt order.
Write $C_1,\ldots,C_K$ for the $K$ cached chunks in prompt order. We compare two executions: a \emph{full-recomputation run} that recomputes all positions, and a \emph{full-reuse run} that loads each chunk $C_k$'s K/V from offline storage produced from $X_{\text{warm},k} = W \sqcup C_k$, where $W \subseteq \Pseg$ is a shared left prefix that may be empty.
We index transformer layers by $\ell = 0, 1, \dots, L_{\text{pref}}-1$, where $L_{\text{pref}}$ is the number of transformer blocks. We use $\ell = 0$ to denote the K/V projections of the layer-$0$ input after positional alignment. Transformer block $\ell$ produces the hidden states from which layer-$(\ell+1)$ K/V are projected.
Let $V_j^{(\ell),\textsc{full}}$ and $V_j^{(\ell),\textsc{cache}}$ denote the value vectors for token $j$ under full recomputation and offline caching. The per-token value deviation at layer $\ell$ is the vector $\Delta V_j^{(\ell)} = V_j^{(\ell),\textsc{full}} - V_j^{(\ell),\textsc{cache}}$ for $j \in \Cseg$. The key deviation $\Delta K_j^{(\ell)}$ uses the same full-minus-cache convention.
Here, $\Delta K_{\Cseg}^{(\ell)}$ and $\Delta V_{\Cseg}^{(\ell)}$ stack the per-token deviations of all reused tokens, i.e., $\Delta V_{\Cseg}^{(\ell)} = [\Delta V_j^{(\ell)}]_{j \in \Cseg}$.
The cache mismatch over the reused region at layer $\ell$ is
\begin{equation}
\label{eq:runtime_state}
E_{\Cseg}^{(\ell)} \;\coloneqq\; \bigl\|\,[\,\Delta K_{\Cseg}^{(\ell)},\ \Delta V_{\Cseg}^{(\ell)}\,]\,\bigr\|_F,
\end{equation}
where $\|\cdot\|_F$ is the Frobenius norm.
After positional alignment, $E_{\Cseg}^{(0)} = 0$ and cache mismatch on $\Cseg$ can first appear at layer $1$.

\begingroup
\paragraph{Multiple retrieved chunks.}
In a typical prompt $[\text{sys}][C_1]\cdots[C_K][\text{question}]$, the cache of $C_k$ is computed offline under $W \sqcup C_k$, where $W$ typically containing the system prompt. Thus, the reused cache of $C_k$ misses the preceding context $M_k=(\Pseg\setminus W)\cup C_1\cup\cdots\cup C_{k-1}$. We analyze the one chunk $C_k$ setting in our main problem formulation, but the analysis can extend to all chunks, as shown in Section~\ref{sec:theory:synthesis}.
\endgroup

The analysis for this problem formulation in this paper relies on five regularity conditions: causal decoder, position-aligned reused cache, $W$ as a left prefix of $\Pseg$, a per-chunk view of the source mismatch, and local Lipschitz smoothness of layer maps and the output head. Full notation and assumptions are provided in Appendix~\ref{app:setup}.

\subsection{Cached-Chunk Mismatch on the Reused Region}
\label{sec:theory:source}

Cached-chunk mismatch arises at a token $i \in C_k$ when, under the full-recomputation run, it would have allocated substantial attention mass to the missing preceding context $M_k$, and that context carries information non-redundant with what is visible in the warmup.

\begin{proposition}[Cached-chunk mismatch identity]
\label{prop:source}
At layer $0$, consider one attention head before its output projection. Let $e_i^{(0)}$ be its output under full recomputation minus its output under offline caching at token $i$. Let $\alpha_i^{(0)}$ be the total attention weight assigned to $M_k$ under full recomputation. The vectors $\mu_{\text{miss},i}^{(0)}$ and $\mu_{\text{vis},i}^{(0)}$ are attention-weighted value means over $M_k$ and the context visible offline, respectively, with weights normalized within each set. Define their gap as $\Delta\mu_i^{(0)}=\|\mu_{\text{miss},i}^{(0)}-\mu_{\text{vis},i}^{(0)}\|_2$, where $\|\cdot\|_2$ is the Euclidean norm. Then the per-token attention-output mismatch on $\Cseg$ admits the exact identity
\begin{equation}
\label{eq:source_identity}
e_i^{(0)} \;=\; \alpha_i^{(0)}\bigl(\mu_{\text{miss},i}^{(0)} - \mu_{\text{vis},i}^{(0)}\bigr),
\qquad
\bigl\|e_i^{(0)}\bigr\|_2 \;=\; \alpha_i^{(0)}\,\Delta\mu_i^{(0)}.
\end{equation}
\end{proposition}

Splitting the full attention output into visible- and missing-context averages gives the identity: offline attention retains only the visible-context average.

The layer-$0$ identity is exact under position alignment. Since the layer-$0$ K/V of $\Cseg$ coincide in the two runs with $E_{\Cseg}^{(0)}=0$, the mismatch injected at layer $0$ can first appear in the layer-$1$ cache. Layer $1$ is therefore the earliest layer that can carry a measurable signal, which motivates the check layer in Section~\ref{sec:method:algo}. Subsequent layers accumulate this mismatch through a layerwise recursion derived in Appendix~\ref{app:proof:source:recursion}.

% --- 3.3 Suffix-side propagation ---
\subsection{Suffix-Side Propagation through Prefill Attention}
\label{sec:theory:suffix}
\begingroup
Although the suffix $\Sseg$ is computed online, its queries attend to the keys and values of $\Cseg$, which can differ between the two runs from layer $1$ onward (Section~\ref{sec:theory:source}). The cache mismatch therefore perturbs the hidden states of $\Sseg$, from which decoding starts; we call this perturbation the \emph{suffix-side drift}. This subsection bounds the drift injected into a suffix token $u$ at a single layer and shows that each reused token $j$ enters it weighted by $a_{uj}$ (\textit{i.e.,} the attention that $u$ assigns to $j$).
\endgroup

% Errors on $\Cseg$ do not stay on $\Cseg$.
% At every layer $\ell \ge 1$, suffix tokens attend to the reused cache. The bound below relates the resulting drift to the cache mismatch and the attention it receives.

\begingroup
For one attention head at a fixed layer, let $q_u$ be the query vector of suffix token $u$, and let $K_j$ and $V_j$ be the key and value vectors with a dimension $d$ at position $j$. The indices $j$ and $t$ range over the causally visible positions in $\mathcal T_u$, including $u$ itself. The attention weight $a_{uj}$ and output $o_u$ are
\begin{equation}
\label{eq:attention_reference}
a_{uj}=\frac{\exp(q_u^\top K_j/\sqrt d)}{\sum_{t\in\mathcal T_u}\exp(q_u^\top K_t/\sqrt d)},
\qquad o_u=\sum_{j\in\mathcal T_u}a_{uj}V_j,
\end{equation}
We temporarily omit the layer index and evaluate these quantities under full recomputation.
To isolate the effect of the cached KV, we hold the query and the KV outside $\Cseg$ fixed at their full-recomputation values. Let $\delta o_{u,\Cseg}$ denote the output computed with the full-recomputation KV on $\Cseg$ minus the output computed with the cached KV on $\Cseg$. A first-order expansion around full recomputation gives
\begin{equation}
\label{eq:suffix_expansion}
\delta o_{u,\Cseg}
=\underbrace{\sum_{j\in\Cseg}a_{uj}\Delta V_j}_{\text{value term}}
+\underbrace{\sum_{j\in\Cseg}a_{uj}\frac{q_u^\top\Delta K_j}{\sqrt d}(V_j-o_u)}_{\text{key term}}
+O\!\left(E_{\Cseg}^2\right).
\end{equation}
The remainder $O(E_{\Cseg}^2)$ collects second- and higher-order terms in the cache mismatch. The value term changes the content read at fixed attention weights. The key term changes those weights and redistributes attention relative to the mean output $o_u$.
Both terms carry the attention factor $a_{uj}$, but the key term also depends on the direction of $\Delta K_j$ relative to $q_u$ and on $V_j-o_u$.
\endgroup

\begin{proposition}[Suffix drift through prefill attention]
\label{prop:suffix}
For a suffix token $u \in \Sseg$ and a layer $\ell \ge 1$, the attention-output perturbation contributed by $\Cseg$ admits, at first order in $E_{\Cseg}^{(\ell)}$,
\begin{equation}
\label{eq:suffix_direct}
\bigl\|\delta o_{u,\Cseg}^{(\ell)}\bigr\|_2 \;\le\; G_{u,\Cseg}^{(\ell)}\,E_{\Cseg}^{(\ell)} \;+\; O\!\bigl((E_{\Cseg}^{(\ell)})^2\bigr),
\end{equation}
where $G_{u,\Cseg}^{(\ell)}$ is a nonnegative sensitivity factor determined by the reference attention weights, query norm and value differences $V_j-o_u$. Its full expression and derivation are given in Appendix~\ref{app:proof:suffix:pertok}.
Aggregating across the suffix and unrolling across layers gives an aggregate bound on suffix-side drift, derived in Appendix~\ref{app:proof:suffix:aggregate}.
\end{proposition}

% Eq.~\eqref{eq:suffix_direct} provides an upper bound in terms of the aggregate cache mismatch: the bound depends on the suffix-to-$\Cseg$ attention $L^2$ norm and the value differences from the mean attention output.
% For the \emph{token selection}, KVFocus operates at a finer granularity: Section~\ref{sec:theory:synthesis} carries out a per-token decomposition of the first-order perturbation that the aggregate bound aggregates away.

The suffix-side drift also perturbs the suffix KV that the decoding steps read. Under greedy decoding, the resulting logit perturbation is bounded by the KV mismatch on all preceding positions, and a decoded token remains unchanged as long as this perturbation is below half of the greedy margin. Please see Proposition~\ref{prop:decode} in Appendix~\ref{app:proof:decode} for the full derivation. Note that we use this result only to justify reducing the drift. None of its quantities is used to compute the token-risk score.

% Proposition~\ref{prop:decode} in Appendix~\ref{app:proof:decode} connects prompt-side drift to greedy-decoding stability, although no quantity from it enters the score or the selector.

% --- 3.5 Synthesis ---
\subsection{Per-Token First-Order Decomposition of the Value Term}
\label{sec:theory:synthesis}
Proposition~\ref{prop:suffix} bounds the perturbation by the aggregate cache mismatch $E_{\Cseg}^{(\ell)}$. This shows that the mismatch reaches the suffix, but it does not indicate which reused tokens should be recomputed. Since the expansion in Eq.~\eqref{eq:suffix_expansion} is a sum over $j \in \Cseg$, we instead examine the contribution of each reused token. We focus on the value term, which can be measured at the check layer. The key term induces a similar token ranking, and adding it to the score does not significantly change F1 (Appendix~\ref{app:keyterm}).

% We focus on the value term, which can be measured at the check layer and yields an efficient token-ranking signal. The reason of not using the key term is discussed in Appendix~\ref{app:keyterm}.}

% To identify how each reused token contributes to the local suffix-side perturbation, we decompose the suffix-side perturbation $\delta o_{u,\Cseg}^{(\ell)}$ in Proposition~\ref{prop:suffix} into per-token contributions before aggregation.

% KVFocus uses the V-term because it is directly observable at the check layer and yields an efficient token-ranking signal. Appendix~\ref{app:keyterm} shows that adding the key-drift term leaves most of the selection unchanged and does not significantly change F1 at the evaluated budgets.

For a fixed suffix token $u$ and attention head, write $\delta o_{u,\Cseg}^{(\ell),V}$ for the value component of the local perturbation. The V-term is the attention-weighted sum of per-token value deviations,
\begin{equation}
\label{eq:per_token_decomp}
\delta o_{u,\Cseg}^{(\ell),V} \;=\; \sum_{j \in \Cseg}\, a_{uj}^{(\ell)} \cdot \Delta V_j^{(\ell)},
\end{equation}
where $a_{uj}^{(\ell)}$ is the attention weight from suffix token $u$ to token $j$ at layer $\ell$, and the head index is suppressed for readability.
Token $j$'s V-term contribution therefore has magnitude
\begin{equation}
\label{eq:per_token_contrib}
c_{uj}^{(\ell)} \;\coloneqq\; a_{uj}^{(\ell)} \cdot \|\Delta V_j^{(\ell)}\|_2.
\end{equation}
Each per-token contribution multiplies a \emph{cache-side} term $\|\Delta V_j^{(\ell)}\|_2$ and a \emph{downstream} term $a_{uj}^{(\ell)}$ that measures the suffix-to-$j$ attention weight. Both factors are local to token $j$ and admit direct runtime measurement at a single check layer.

Eq.~\eqref{eq:per_token_contrib} is stated for a single suffix token and a single head. Appendix~\ref{app:proof:suffix:vterm} averages it over the suffix probes and attention heads, and shows that the averaged contribution of token $j$ is bounded by $m_j U_j$ (the score defined in Section~\ref{sec:method:score}).

Since the RAG chunks occupy disjoint positions, the layer-$0$ identity in Eq.~\eqref{eq:source_identity} and the per-token decomposition in Eq.~\eqref{eq:per_token_decomp} apply to each chunk separately, and the contributions of different chunks to the local suffix-side drift add up at first order around a common reference as detailed in Appendix~\ref{app:setup:multiseg}. The analysis therefore extends to any number of chunks, and KVFocus selects the top-$r\%$ of all reused tokens under one global budget, where $r\%$ is the fraction selected for recomputation.

% Appendix~\ref{app:proof:suffix:vterm} derives the aggregation over suffix probes and attention heads.

% =================================================================
% §4 Method
% =================================================================
\section{Selective Recomputation with KVFocus}
\label{sec:method}

\begin{figure}[!t]
\centering
\includegraphics[width=0.95\linewidth]{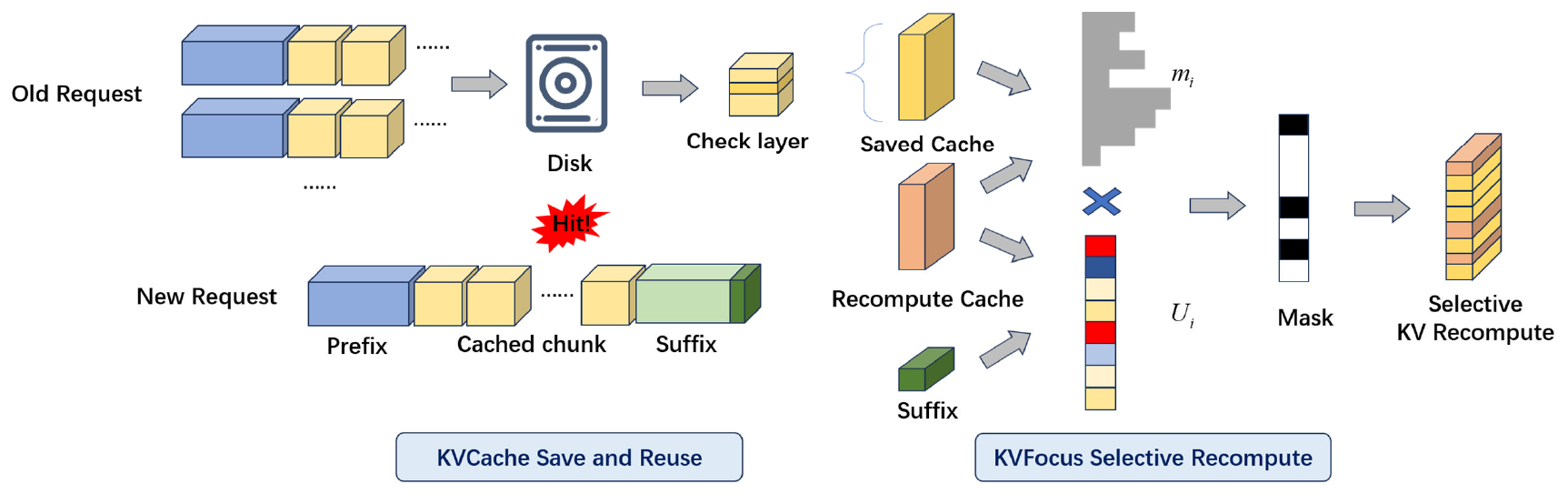}
\caption{System overview. (\textbf{Left}) K/V tensors of retrieved document chunks are computed during prior requests and persisted to disk. On a new request, cached chunks are loaded from disk and inserted between the online prefix and online suffix. (\textbf{Right}) At the check layer, the per-token value deviation $m_i$ and the suffix-probe attention mass $U_i$ (downstream-side) are read out. Their product $s_i = m_i \cdot U_i$ ranks tokens, and the top-$r\%$ are recomputed in the subsequent layers of the same prefill pass, while the rest continue to reuse the loaded cache.}
\label{fig:system}
\end{figure}

This section turns the per-token decomposition in Section~\ref{sec:theory:synthesis} into a selective recomputation method (Fig.~\ref{fig:system}), which ranks the reused tokens by the following per-token risk score:
\begin{equation}
\label{eq:s_def}
s_i \;\coloneqq\; m_i \cdot U_i,
\end{equation}
where $m_i$ is the per-token value deviation at the check layer (cache side) and $U_i$ is the head-averaged suffix-probe attention onto position $i$ (downstream side). 

% The score is used only for ranking and is not calibrated to the magnitude of the answer error.

% KVFocus selects the top-$r\%$ of of reused tokens by $s_i$ for online recomputation, with $r$ a deployment hyperparameter.

% --- 4.1 Score definition ---
\subsection{Per-Token Value Deviation and Attention Mass}
\label{sec:method:score}

% \paragraph{Per-token value deviation.}
For $i \in \Cseg$, the layer-$\lcheck$ value deviation is the $L_2$ distance between the V tensor produced by the full-recompute pass at layer $\lcheck$ and the V tensor loaded from the offline cache, computed over the concatenation of all KV heads:
\begin{equation}
\label{eq:m_def}
m_i \;\coloneqq\; \big\| \bm{V}_i^{(\lcheck),\,\textsc{fc}} \;-\; \bm{V}_i^{(\lcheck),\,\text{cache}} \big\|_2,
\end{equation}
where $\bm{V}_i^{(\lcheck),\,\cdot} \in \mathbb{R}^{H_{kv} d}$ concatenates the per-head V vectors at token $i$ ($H_{kv} \le H_q$ in grouped-query attention). % author-approved 2026-10-15: align eq:m_def with implementation (concatenated single L2; two versions' rank corr rho=0.9995, zero number changes)
$m_i$ is the runtime measurement of the per-token value deviation $\Delta V_j^{(\lcheck)}$ from \eqref{eq:per_token_decomp}; it requires no additional attention computation because the full-recompute V tensors are materialized for the standard layer-$\lcheck$ pass.

% \paragraph{Per-token attention mass.}
For each reused token $i$, the layer-$\lcheck$ suffix-probe-to-$i$ attention is aggregated linearly over probe positions and query heads into a per-token scalar attention mass
\begin{equation}
\label{eq:U_def}
U_i \;\coloneqq\; \frac{1}{|\Spsuff|\,H_q} \sum_{u \in \Spsuff} \sum_{h=1}^{H_q} a_{ui}^{(\lcheck),\,h},
\end{equation}
where $a_{ui}^{(\lcheck),\,h} \in [0, 1]$ is the standard softmax-normalized attention score from probe $u \in \Spsuff$ to key position $i \in \Cseg$ at layer $\lcheck$ and query head $h$, computed against $K_{\Cseg}^{(\lcheck),\,\textsc{fc}}$, and $H_q$ is the number of query heads.
$U_i$ corresponds to the linear coefficient $a_{u,j}$ in the per-token contribution \eqref{eq:per_token_contrib}, aggregated over the suffix probe and heads.

% \paragraph{Suffix probe.}
The suffix probe $\Spsuff \subseteq \Sseg$ is the last $K_p^{\text{suff}}$ online suffix tokens before the first decoded position. The final prefill position directly produces the initial next-token logits, while nearby bridge tokens are visible to the earliest decode queries. We therefore use the last $K_p^{\text{suff}}=5$ suffix tokens as the probe for all three models (Appendix~\ref{app:impl:bridge}).

% --- 4.2 Algorithm ---
\subsection{The KVFocus Selector}
\label{sec:method:algo}
Algorithm~\ref{alg:kvfocus} summarizes KVFocus, which performs token selection within a single prefill pass. Layers $0$ to $\lcheck$ run on all positions as in full recomputation, and at the check layer KVFocus computes $m_i$, $U_i$, and $s_i$ for every reused token. It then selects the top $\lceil r \cdot L_{\Cseg} \rceil$ tokens over all reused chunks, so that the budget is shared across chunks rather than split per chunk. In each layer above $\lcheck$, the selected tokens are recomputed online, whereas the other reused tokens skip computation and contribute their cached K/V to attention. 

% The rest of this subsection explains the choice of $\lcheck$ and why the score uses only the value term.

\begin{algorithm}[!t]
\caption{KVFocus token selection at recomputation rate $r$}
\label{alg:kvfocus}
\begin{algorithmic}[1]
\State Run layers $0$ through $\lcheck$ once on all positions of $X$.
\For{$i \in \Cseg$}
    \State $m_i \gets$ $L_2$ distance between the concatenated $V_i^{(\lcheck),\textsc{fc}}$ and the loaded $V_i^{(\lcheck),\text{cache}}$, Eq.~\eqref{eq:m_def}.
    \State $U_i \gets$ head- and probe-averaged suffix-probe attention onto $i$, Eq.~\eqref{eq:U_def}.
    \State $s_i \gets m_i \cdot U_i$.
\EndFor
\State Select $\mathcal{R} \gets$ top $\lceil r \cdot L_{\Cseg} \rceil$ tokens by descending $s_i$ across all reused chunks.
\State Continue the same forward pass without repeating earlier layers.
\For{layer $\ell = \lcheck + 1, \dots, L_{\text{pref}} - 1$}
    \State Recompute K and V online for tokens in $\mathcal{R}$; load cached K and V for tokens in $\Cseg \setminus \mathcal{R}$.
    \State Run online attention and MLP for $\Pseg \cup \mathcal{R} \cup \Sseg$.
\EndFor
\end{algorithmic}
\end{algorithm}

By position alignment, the layer-$0$ K/V on $\Cseg$ are identical between the full-recomputation and full-reuse runs, so a comparison at layer $0$ yields no signal. Layer $1$ is the earliest layer at which the mismatch injected at layer $0$ can appear in the cache (Section~\ref{sec:theory:source}). We choose $\lcheck = 1$ since it minimizes the cost of layers $0$ to $\lcheck$, which run on all positions before selection.
A deeper check layer may estimate the attention mass more accurately, but it requires more layers to run on all positions, which reduces the latency saving. CacheBlend~\citep{yao2025cacheblend} likewise selects tokens at an early layer.

% \paragraph{Choice of $\lcheck$.}
% By position alignment, layer $0$ K/V on $\Cseg$ are identical between full and full-reuse runs, so any layer-$0$ comparison reveals nothing.
% Layer $1$ is the earliest layer at which the layer-$0$ source injection has propagated into a measurable cache discrepancy (Section~\ref{sec:theory:source}). We choose $\lcheck = 1$ since it minimizes the unconditional layer-$0$-to-$\lcheck$ forward cost.
% A deeper check layer may estimate the attention mass more accurately, but all layers up to it must run on all positions before selection, which reduces the latency saving. Moreover, CacheBlend~\citep{yao2025cacheblend} also selects tokens at the first layer.

% The first-order expansion of $\delta o_{u,\Cseg}^{(\ell)}$ contains the V-term \eqref{eq:per_token_decomp} and a key-drift term that perturbs attention weights through $\Delta K_j^{(\ell)}$.
% We compute $s_i$ from the V-term directly because $\Delta V_j^{(\ell)}$ is observable during the standard layer-$\lcheck$ pass at no additional cost, whereas the key-drift term also involves the query--key-deviation projections and centered value vectors in Eq.~\eqref{eq:suffix_expansion}. In our offline analysis (Appendix~\ref{app:keyterm}), the two terms induce similar token rankings (Spearman $\rho = 0.809$), and adding the key-drift term back into the score does not significantly change F1 ($+0.001$ on average).

% --- 4.3 Cost ---
\subsection{Computational Structure and Implementation}
\label{sec:method:cost}

KVFocus runs a single prefill pass in which no layer is computed twice. As in full recomputation, layers $0$ to $\lcheck$ run on all positions. Above $\lcheck$, the unselected $1-r$ fraction of $\Cseg$ tokens skips per-layer projection, attention, and MLP and uses the cached K/V. Hence, the layer computation of KVFocus never exceeds that of full recomputation. Its extra costs lie outside the layer stack: loading the cached K/V (shared by all reuse methods) and the check-layer scoring, where the value deviation $m_i$ is a memory-side comparison and the suffix-probe-to-token attention statistics for $U_i$ are read or recomputed as a small probe-attention block. All these costs are included in the reported TTFT.
A simple cost model is
\begin{equation}
\label{eq:cost_model}
\frac{\textsf{Cost}_{\text{KVFocus}}}{\textsf{Cost}_{\text{full}}}
\;\approx\;
\frac{\lcheck + 1}{L_{\text{pref}}}
\;+\;
\frac{L_{\text{pref}} - \lcheck - 1}{L_{\text{pref}}}
\cdot
\frac{|\Pseg| + |\Sseg| + rL_{\Cseg}}{|X|}
\;+\; \epsilon_{\text{IO}} + \epsilon_{\text{probe}},
\end{equation}
where $\epsilon_{\text{IO}}$ collects the cache-load overhead and $\epsilon_{\text{probe}}$ the probe-attention readout. We report wall-clock measurements rather than relying on this idealized model (Section~\ref{sec:exp_efficiency}).
% The savings come from skipping computation on unselected cached tokens in the layers above $\lcheck$.
% On short prompts, these fixed costs are not amortized, so KVFocus, like other reuse methods, can be slower than full recomputation (Section~\ref{sec:exp_efficiency}).

\section{Experiments}
\label{sec:experiments}
\label{sec:exp_setup}% DRAFT fix: submission referenced this label but had it commented out (A7 broken-ref commitment)

% \subsection{Setup}
% \label{sec:exp_setup}

We evaluate three open-weight 7--8B instruction-tuned LLMs (Mistral-7B-Instruct-v0.2~\citep{jiang2023mistral7b}, Qwen2.5-7B-Instruct~\citep{qwen2025qwen25technicalreport}, Qwen3-8B~\citep{yang2025qwen3}) on seven datasets covering three task types: four multi-hop RAG-QA datasets (2WikiMQA~\citep{ho2020constructing}, HotpotQA~\citep{yang2018hotpotqa}, MuSiQue~\citep{trivedi2022musique}, MultiHop-RAG~\citep{tang2024multihop}), a position-sensitive stress test (Needle-in-a-Haystack~\citep{kamradt2023needle}), and two non-QA generation tasks (SAMSum~\citep{gliwa2019samsum}, Multi-News~\citep{fabbri2019multi}). We implement KVFocus on top of the SGLang serving framework~\citep{zheng2024sglang}, and provide implementation details in Appendix~\ref{app:impl}. All runs use greedy decoding on a single \textsc{NVIDIA H200} GPU with FP16 weights and batch size $1$. Baselines are Full Recomputation (FR), Naive Full Reuse, EPIC~\citep{hu2024epic}, CacheBlend~\citep{yao2025cacheblend}, and ProphetKV~\citep{wang2026prophetkv}, which are all run on the same SGLang stack at six matched actual recomputation rates $r \in \{1, 2, 5, 10, 15, 20\}\%$. Appendix~\ref{app:related_landscape} summarizes how each baseline selects tokens.

We report two answer-quality metrics: LongBench~\citep{bai2024longbench} F1 across all $7\!\times\!3\!\times\!6 = 126$ cells, where \emph{a cell denotes a (model, dataset, rate) triple}, and an automated LLM-as-judge score using \texttt{Qwen3.5-397B-A17B}~\citep{qwen35moe} that scores each prediction against the gold answer on a five-level rubric $\{0, 0.25, 0.5, 0.75, 1.0\}$ (full prompt in Appendix~\ref{app:impl:judge}). Judge scores cover $108$ cells, excluding MultiHop-RAG, whose $2{,}250$ samples per cell exceeded the judge compute budget. For each baseline, the matched-rate F1 gap $\Delta\mathrm{F1}$ is computed by averaging paired F1 differences (KVFocus minus baseline) within each cell and then averaging uniformly over cells. $\Delta\mathrm{Judge}$ is computed analogously. We use three efficiency metrics: \emph{TTFT ratio} $= \mathrm{TTFT}^{(b)} / \mathrm{TTFT}^{(\mathrm{KVFocus})}$, \emph{speedup vs.\ FR} $= \mathrm{TTFT}^{(\mathrm{FR})} / \mathrm{TTFT}^{(\mathrm{method})}$ (\textbf{Spd}), and \emph{quality retention} $= \overline{\mathrm{F1}}^{(\mathrm{method})} / \overline{\mathrm{F1}}^{(\mathrm{FR})}$ (\textbf{Ret}). Following CacheBlend~\citep{yao2025cacheblend} and EPIC~\citep{hu2024epic}, we first average TTFT over all test requests and then take the ratio of the averages, rather than averaging per-request ratios. Bridge tokens, hardware stack, and ProphetKV TTFT accounting are detailed in Appendix~\ref{app:impl}.

\subsection{Comparison with Selective-Recomputation Baselines}

% \subsubsection{Quality Comparison at Matched Recomputation Rates and Matched TTFT Budgets}
\label{sec:exp_matched_rate}
\begin{table}[!t]
\caption{KVFocus vs.\ baselines: cell-level macro-averaged $\Delta\mathrm{F1}$ over 126 cells and $\Delta\mathrm{Judge}$ over 108 cells (excl.\ MultiHop-RAG), where we report 95\% CI (confidence intervals) to quantify the uncertainty. Wins is the count of cells where KVFocus exceeds the baseline. TTFT ratio is averaged over 126 cells. A value below $1\times$ means that KVFocus is slower than the baseline.}
\label{tab:matched_rate_summary}
\centering
\small
\setlength{\tabcolsep}{4pt}
\begin{tabular}{lccccccc}
\toprule
                & \multicolumn{2}{c}{$\Delta$F1} & \multicolumn{2}{c}{$\Delta$Judge} & TTFT  & \multicolumn{2}{c}{Wins} \\
\cmidrule(lr){2-3} \cmidrule(lr){4-5}\cmidrule(lr){7-8}
Baseline        & Mean & 95\% CI & Mean & 95\% CI & ratio & /126 & /108 \\
\midrule
% values: submitted version (restored)
CacheBlend      & $\textcolor{black}{+0.081}$ & $\textcolor{black}{[+0.062,+0.103]}$ & $\textcolor{black}{+0.162}$ & $\textcolor{black}{[+0.120,+0.206]}$ & $\textcolor{black}{0.98\times}$ & $\textcolor{black}{109}$ & $\textcolor{black}{85}$ \\
EPIC            & $\textcolor{black}{+0.044}$ & $\textcolor{black}{[+0.029,+0.061]}$ & $\textcolor{black}{+0.095}$ & $\textcolor{black}{[+0.063,+0.129]}$ & $\textcolor{black}{0.62\times}$ & $\textcolor{black}{93}$ & $\textcolor{black}{82}$ \\
ProphetKV       & $-0.016$ & $[-0.030,-0.002]$ & $-0.019$ & $[-0.036,+0.002]$ & $\textcolor{black}{1.71\times}$ & $46$ & $43$ \\
\bottomrule
\end{tabular}
\end{table}

% \paragraph{Quality gains over CacheBlend and EPIC.}
Table~\ref{tab:matched_rate_summary} illustrates the quality comparison at matched recomputation rates. KVFocus improves F1 over CacheBlend by $+0.081$ ($87\%$ of cells) and over EPIC by $+0.044$ ($74\%$ of cells), with CIs that exclude zero. Under the LLM judge, the same gaps roughly double: $+0.162$ over CacheBlend and $+0.095$ over EPIC. The widening under semantic evaluation indicates that KVFocus's gains include answers that are correct in meaning but penalized by surface-form F1.
% ProphetKV is the strongest baseline. KVFocus still improves over it by $-0.016$ F1 and $-0.019$ Judge, with CIs excluding zero, winning on $36.5\%$ of cells under F1 and $27.8\%$ under Judge---a small but statistically significant edge. The decisive separation is the time axis: KVFocus is on average $2.02\times$ faster than ProphetKV at the same recomputation rate, because ProphetKV requires an additional mask-scoring pass that KVFocus avoids by reusing the check-layer activations already computed for $m_i$. 
ProphetKV is the strongest baseline in average quality: it exceeds KVFocus by $0.016$ F1 and $0.019$ Judge score. The F1 difference is statistically significant, whereas the Judge difference is not, with its $95\%$ confidence interval including zero. At the same recomputation rate, however, ProphetKV's TTFT is on average $1.71\times$ that of KVFocus, because it requires an additional mask-scoring pass, whereas KVFocus performs token selection within the same prefill pass.

% Section~\ref{sec:exp_matched_ttft} compares the two methods at matched TTFT budgets. KVFocus is essentially TTFT-tied with CacheBlend ($0.98\times$) and slower than EPIC ($0.86\times$). The EPIC advantage comes at substantially lower quality (Section~\ref{sec:exp_efficiency}).

\begingroup
% \subsection{Quality under matched TTFT budgets}
% \label{sec:exp_matched_ttft}

\begin{table}[t]

\caption{ Quality under matched TTFT budgets $\mathrm{TTFT}^{\mathrm{FR}}_c/s$. Entries are per-sample paired $\Delta$F1 (KVFocus minus baseline), pooled over the cells where both methods have a tested rate within the budget. $^{*}$ indicates that the corresponding 95\% confidence interval excludes zero. --: no cell has a tested ProphetKV rate within the budget.}
\label{tab:iso_panel}
\centering
\small
\setlength{\tabcolsep}{5pt}
\begin{tabular}{lccccc}
\toprule
 & \multicolumn{5}{c}{Budget $s$} \\
\cmidrule(lr){2-6}
Baseline & $1.25\times$ & $1.5\times$ & $1.75\times$ & $2\times$ & $2.5\times$ \\
\midrule
CacheBlend & $+0.015^{*}$ & $+0.020^{*}$ & $+0.022^{*}$ & $+0.026^{*}$ & $-0.026^{*}$ \\
EPIC & $+0.007^{*}$ & $+0.012^{*}$ & $+0.010^{*}$ & $-0.004$ & $-0.060^{*}$ \\
ProphetKV & $+0.071^{*}$ & -- & -- & -- & -- \\
\bottomrule
\end{tabular}
\end{table}

Because token selection has a different cost in each method, the same recomputation rate $r$ yields different TTFTs. We therefore also compare the methods under matched latency budgets $B(c, s) = \mathrm{TTFT}^{\mathrm{FR}}_c / s$, where $c$ is a model-dataset combination and $s \in \{1.25, 1.5, 1.75, 2, 2.5\}$ is the target speedup over FR. For each $c$ and $s$, each method uses the largest tested rate whose TTFT fits the budget, chosen without consulting F1, and is infeasible if no tested rate fits. Table~\ref{tab:iso_panel} reports the paired F1 differences pooled over the combinations where both methods are feasible.

The quality advantage of ProphetKV at a fixed rate does not carry over to matched budgets. ProphetKV meets only the loosest budget ($1.25\times$), where KVFocus improves F1 by $0.071$. For $s \ge 1.5$, its mask-scoring pass leaves no tested rate within the budget in any combination, whereas KVFocus remains feasible. KVFocus also significantly outperforms CacheBlend and EPIC at every budget up to $2\times$, except for a non-significant difference from EPIC at $2\times$. Only at the tightest budget ($2.5\times$) does KVFocus fall behind both, likely because it is slower than EPIC and marginally slower than CacheBlend at a fixed rate (Table~\ref{tab:matched_rate_summary}), which leaves it lower feasible rates under a tight budget.

\endgroup%

% \subsubsection{Quality-TTFT Trade-Off}
% \label{sec:exp_pareto}

\begin{figure}[!t]
  \centering
  \begin{subfigure}[b]{0.42\textwidth}
    \centering
    \includegraphics[width=\linewidth]{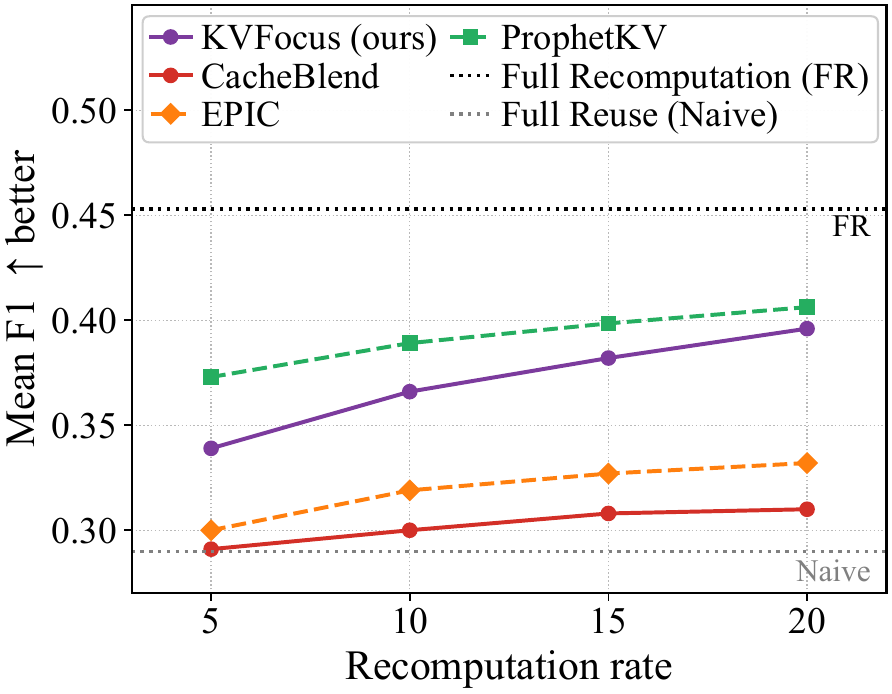}
    \caption{Mean F1 vs.\ recomputation rate $r$}
    \label{fig:rate}
  \end{subfigure}
  \hspace{0.04\textwidth}%
  \begin{subfigure}[b]{0.42\textwidth}
    \centering
    \includegraphics[width=\linewidth]{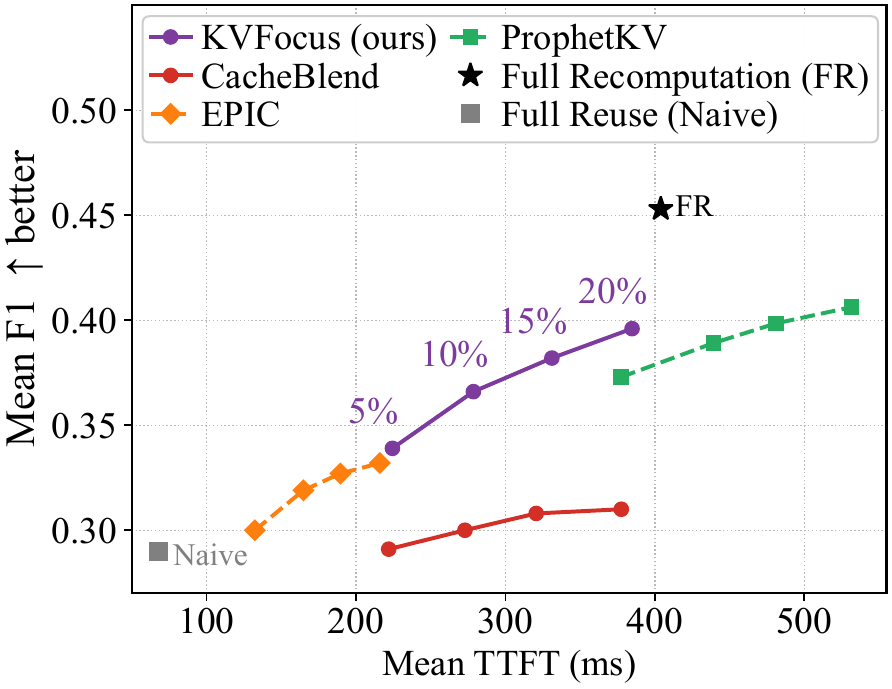}
    \caption{Mean F1 vs. mean TTFT.}
    \label{fig:ttft}
  \end{subfigure}
  \caption{Performance comparison in terms of mean F1, recomputation rate, and TTFT.}
  \label{fig:results}
\end{figure}

\paragraph{Quality-TTFT trade-off.}
Fig.~\ref{fig:results} plots mean F1 against recomputation rate and mean F1 against TTFT, averaged over $21$ model-dataset cells ($3$ models $\times$ $7$ datasets) at each rate. KVFocus consistently improves mean F1 over CacheBlend and EPIC across the tested rates $r$, with larger gains at higher rates. ProphetKV achieves higher mean F1, but incurs substantially higher TTFT. The TTFT curves match the relationships in Table~\ref{tab:matched_rate_summary}: KVFocus is marginally slower than CacheBlend, EPIC is faster at lower average F1, and ProphetKV is consistently the slowest selector. KVFocus therefore combines stronger quality than CacheBlend and EPIC with lower latency than ProphetKV.

% Student's subsection restored verbatim (PI, 2026-10-26). TODO(student): the data come from the RTX 5090 re-run
% and must be updated to match the rest of the paper.
% ---------------------------------------------------------------------------
% NEW subsection: matched TTFT budgets (blue). Values = isospd22 canon keys
% (isospd22_S_minus_CB_s1p5_pooled etc.); wire via cr_numbers.tex in final pass.
% ---------------------------------------------------------------------------

\paragraph{Robustness across datasets.}
% \label{sec:exp_per_dataset}

KVFocus improves F1 over CacheBlend across all seven datasets, with its largest gains over CacheBlend and EPIC on HotpotQA, SAMSum, and Needle-in-a-Haystack. ProphetKV remains stronger on the four multi-hop QA datasets, while KVFocus has small positive gains on the remaining tasks (Appendix~\ref{app:detailed_tables}). The Mistral Needle case at $r{=}10\%$ highlights the importance of token selection: CacheBlend and EPIC both fall below Naive Full Reuse, whereas KVFocus raises F1 from $0.264$ to $0.588$, exceeding ProphetKV's $0.528$. This pattern is consistent with the benefit of directing recomputation toward tokens that are both deviated and attended by the suffix. On Multi-News, the negligible FR$-$Naive headroom leaves little room for improvement, and all methods achieve nearly identical F1.

\subsection{Comparison with Full Recomputation}
\label{sec:exp_efficiency}

Table~\ref{tab:efficiency} summarizes the quality-TTFT trade-off relative to FR using mean speedup, F1 retention, and the \emph{headroom recovery}, \emph{i.e.,} the fraction of the FR$-$Naive headroom that a method recovers:
\[
(\overline{\mathrm{F1}}^{(a)}_c - \overline{\mathrm{F1}}^{\mathrm{Naive}}_c) \,/\, (\overline{\mathrm{F1}}^{\mathrm{FR}}_c - \overline{\mathrm{F1}}^{\mathrm{Naive}}_c),
\]
averaged over the $108$ cells whose FR$-$Naive headroom is at least $0.005$ (headroom recovery is undefined for the others). Full per-rate breakdowns including $r \in \{2, 15\}\%$ are in Appendix~\ref{app:throughput}.

\begin{table}[!t]
\caption{Speedup vs.\ FR (\textbf{Spd}, ratio of mean TTFTs over the $21$ cells at each rate), quality retention (\textbf{Ret}, $\overline{\mathrm{F1}}^{(a)}/\overline{\mathrm{F1}}^{\mathrm{FR}}$), and headroom recovery over the $108$ cells with FR$-$Naive headroom $\geq 0.005$. At the same $r$, KVFocus is marginally slower than CacheBlend. Its TTFT advantage at matched quality comes from needing a lower recomputation rate.}
\label{tab:efficiency}
\centering
\small
\setlength{\tabcolsep}{3pt}
\begin{tabular}{l|cc|cc|cc|cc|c}
\toprule
       & \multicolumn{2}{c|}{$r{=}1\%$} & \multicolumn{2}{c|}{$r{=}5\%$} & \multicolumn{2}{c|}{$r{=}10\%$} & \multicolumn{2}{c|}{$r{=}20\%$} & Headroom \\
       & Spd & Ret & Spd & Ret & Spd & Ret & Spd & Ret & recovery \\
\midrule
% values: submitted version (restored)
\textbf{KVFocus} & $\textcolor{black}{\mathbf{2.22\times}}$ & $\textcolor{black}{\mathbf{0.82}}$ & $\textcolor{black}{\mathbf{1.80\times}}$ & $\textcolor{black}{\mathbf{0.86}}$ & $\textcolor{black}{\mathbf{1.45\times}}$ & $\textcolor{black}{\mathbf{0.89}}$ & $\textcolor{black}{\mathbf{1.05\times}}$ & $\textcolor{black}{\mathbf{0.93}}$ & $\textcolor{black}{\mathbf{+36.4\%}}$ \\
ProphetKV  & $\textcolor{black}{1.21\times}$ & $0.79$ & $\textcolor{black}{1.07\times}$ & $0.89$ & $\textcolor{black}{0.92\times}$ & $0.93$ & $\textcolor{black}{0.76\times}$ & $0.97$ & $+44.6\%$ \\
EPIC       & $\textcolor{black}{3.77\times}$ & $\textcolor{black}{0.76}$ & $\textcolor{black}{3.05\times}$ & $\textcolor{black}{0.77}$ & $\textcolor{black}{2.45\times}$ & $\textcolor{black}{0.80}$ & $\textcolor{black}{1.87\times}$ & $\textcolor{black}{0.82}$ & $\textcolor{black}{+16.1\%}$ \\
CacheBlend & $\textcolor{black}{2.24\times}$ & $\textcolor{black}{0.71}$ & $\textcolor{black}{1.82\times}$ & $\textcolor{black}{0.71}$ & $\textcolor{black}{1.48\times}$ & $\textcolor{black}{0.72}$ & $\textcolor{black}{1.07\times}$ & $\textcolor{black}{0.76}$ & $+24.8\%$ \\
\bottomrule
\end{tabular}
\end{table}

The results in Table~\ref{tab:efficiency} indicate that KVFocus achieves an average $2.22\times$ TTFT speedup (per-cell peak speedup reaches $4.37\times$ on long-context multi-hop QA in Appendix~\ref{app:throughput:peak}) only with an $18\%$ quality degradation in comparison to the full recomputation method using a recomputation rate of $r{=}1\%$. Compared to the baselines, KVFocus achieves a better quality-latency trade-off. For example, KVFocus performs better than CacheBlend at all tested recomputation rates, with higher F1 retention and comparable speedup. ProphetKV achieves higher retention at $r \geq 5\%$ ($0.97$ at $r{=}20\%$), but its speedup drops below $1\times$ for $r \geq 10\%$. EPIC achieves the largest raw speedup (up to $3.77\times$ at $r{=}1\%$) at consistently lower retention.

On average, KVFocus recovers $36.4\%$ of the FR$-$Naive headroom, compared with $44.6\%$ for ProphetKV, $24.8\%$ for CacheBlend, and $16.1\%$ for EPIC. KVFocus thus recovers substantially more headroom than CacheBlend at comparable speedups. This is consistent with our analysis: ranking tokens without the suffix attention mass spends budget on tokens whose recomputation does not reach the answer. ProphetKV recovers more, but at a speedup of at most $1.21\times$.

At a fixed recomputation rate, KVFocus is marginally slower than CacheBlend (a TTFT ratio of $0.98\times$ in Table~\ref{tab:matched_rate_summary}) and slower than EPIC, which does not compute token-selection scores. However, KVFocus achieves higher quality with a smaller recomputation budget. Within the tested range, CacheBlend reaches a maximum F1 retention of $0.76$ at $r{=}20\%$, below KVFocus's $0.82$ at $r{=}1\%$. EPIC first reaches the same retention of $0.82$ at $r{=}15\%$ (Table~\ref{tab:efficiency_full}), with a $2.13\times$ speedup over FR, compared with KVFocus's $2.22\times$ at $r{=}1\%$. Thus, KVFocus achieves comparable quality to EPIC with a much smaller recomputation budget and lower latency.

KVFocus does not close the gap to FR: at $r{=}1\%$ it loses $18\%$ of FR's F1. Raising $r$ narrows the gap at the cost of latency, and KVFocus reduces to FR at $r{=}100\%$. Our oracle study on Mistral-7B (Appendix~\ref{app:oracle}) shows that even selectors using last-layer quantities from an FR reference run recover only $52.5\%$, $74.1\%$, and $91.4\%$ of the FR$-$Naive headroom at $r{=}1\%$, $10\%$, and $20\%$, while KVFocus trails these oracles by $9.3$ and $12.2$ points at $r{=}1\%$ and $10\%$. Closing the gap thus requires both a better online proxy and a larger budget. 

Reuse also does not pay off on short inputs: on Multi-News (${\sim}40$\,ms FR TTFT), the TTFT of KVFocus is $1.27$--$1.56\times$ that of FR, and the FR$-$Naive headroom is below $0.005$. Our prompt-length sweep (Appendix~\ref{app:length_sweep}) places the break-even length between about $0.8$K and $1.5$K tokens, and at $256$--$768$ tokens even naive reuse is slower than FR. We therefore position KVFocus as a better operating curve for latency-constrained, long-context RAG serving rather than as a drop-in replacement for full prefill in accuracy-critical deployments.

\subsection{Ablation on Score Composition}
\label{sec:exp_ablation}

We ablate the structure of the score on four representative cells covering both aggregate-evidence (HotpotQA) and position-sensitive (Needle) workloads (full results in Appendix~\ref{app:additional_ablation}). Cache-side deviation alone ($s_j = m_j$, equivalent to CacheBlend's ranking up to constants) improves over Naive on the two HotpotQA cells evaluated with Mistral but \emph{collapses to $0.195$ on Needle-Mistral, below both Random ($0.252$) and Naive ($0.264$)}, because recomputing high-deviation but low-attention tokens spends budget on cache regions the suffix barely attends to. Rank-normalized additive composition $m{+}U$ uses the same two factors as KVFocus but combines them by addition; despite identical inputs, it stays within $\pm 0.012$ F1 of Random on every cell. Among these variants, only the multiplicative form $m{\cdot}U$, which follows from the first-order V-term (Eq.~\eqref{eq:per_token_decomp}), works in both regimes: it beats the strongest of them on all four cells, by $+0.029$ on HotpotQA-Mistral-$r{=}20\%$ to $+0.327$ on Needle. With identical inputs, it also improves over the additive form by $0.031$--$0.327$ F1, so the form of combination matters, and KVFocus is not a heuristic fusion of the CacheBlend and ProphetKV signals. An attention-only variant performs on par with $m{\cdot}U$ on these cells, slightly exceeding it (Appendix~\ref{app:additional_ablation}).

% Among these variants, only the multiplicative form $m{\cdot}U$, the form derived from the first-order V-term (Eq.~\eqref{eq:per_token_decomp}), survives both regimes, winning all four cells with margins over the strongest of these variants from $+0.029$ on HotpotQA-Mistral-$r{=}20\%$ to $+0.327$ on Needle. An attention-only variant is competitive with $m{\cdot}U$ on these four cells (Appendix~\ref{app:additional_ablation}). The form of combination is part of the contribution. This also distinguishes KVFocus from a heuristic fusion of the deviation signal of CacheBlend and the attention signal of ProphetKV: with identical inputs, the additive combination performs close to random selection, whereas the derived multiplicative form improves over it by $0.031$--$0.327$ F1 on these cells.

% ============================================================
% 8. CONCLUSION
% ============================================================
\section{Conclusion}
\label{sec:conclusion}
In this paper, we proposed KVFocus, a selective KV cache reuse method for RAG. A first-order analysis shows that a reused token's contribution to the suffix-side drift is the product of its value deviation and the attention mass it receives from the suffix. KVFocus computes this product as a training-free score at a single check layer and recomputes the top-ranked tokens within the same prefill pass. Across three 7--8B LLMs and seven datasets, it improves the quality-TTFT trade-off over prior selective-recomputation methods and achieves an average $2.22\times$ TTFT speedup over full recomputation at an $18\%$ quality degradation.

\let\section\kvfocusOriginalSection
\let\subsection\kvfocusOriginalSubsection

\begin{ack}
This work was supported in part by the National Natural Science Foundation of China under Grant 62401366, Grant U24A20251, Grant 62320106003, Grant 62595731, Grant 62125109, Grant 62431017, Grant 62371288, in part by the National Key Research and Development Program of China under Grant 2025YFF0515602, in part by Fundamental and Interdisciplinary Disciplines Breakthrough Plan of the Ministry of Education of China (JYB2025XDXM611), and in part by AI for Science Seed Program of Shanghai Jiao Tong University (2025AI4S-QY09).
\end{ack}

% ============================================================
% References (do not count toward 9-page limit)
% ============================================================
\newpage
\bibliographystyle{unsrtnat}
\bibliography{kvfocus_references}

% ============================================================
% Appendix
% ============================================================
\input{appendix_camera_ready_v2.tex}

% ============================================================
% NeurIPS 2026 Paper Checklist (required for submission)
% ============================================================
\newpage

\end{document}

%% file: appendix_camera_ready_v2.tex
% ============================================================
% appendix.tex
% NeurIPS 2026 -- KVFocus
% ============================================================

\appendix
\counterwithin{proposition}{section}
% =================================================================
% A. Notation, Assumptions, and Setup
% =================================================================

\clearpage

\section{Notation, Assumptions, and Multi-Chunk Generalization}
\label{app:setup}

This appendix collects the notation, regularity assumptions, and the multi-chunk generalization referenced from Section~\ref{sec:theory:setup}.

\subsection{Notation}
\label{app:setup:notation}

We use the symbols summarized in Table~\ref{tab:notation}.

\begin{table}[h]
\caption{Notation used throughout the theoretical analysis.}
\label{tab:notation}
\centering
\small
\begin{tabular}{ll}
\toprule
\textbf{Symbol} & \textbf{Meaning} \\
\midrule
$X = P \sqcup C \sqcup S$ & Prompt, online prefix, reused region, online suffix \\
$L_C$ & Token lengths of $C$ \\
$W \subseteq P$ & Shared left-prefix between offline warmup and online context \\
$C=C_1\sqcup\cdots\sqcup C_K$ & Reused chunks in serving order \\
$M_k=(P\setminus W)\cup\bigcup_{b<k}C_b$ & Context missing from the offline encoding of $C_k$ \\
$X_{\text{warm},k} = W \sqcup C_{k}$ & Offline warmup context used to encode chunk $C_k$ \\
$L_{\text{pref}}$ & Number of transformer layers used in prefill \\
$\ell, \ell^\star$ & Layer index and \emph{check layer}, with $\ell^\star = 1$ \\
$H_q, H_{kv}$ & Number of query and key/value heads in grouped-query attention \\
$d$ & Per-head hidden dimension \\
$K_C^{(\ell)}, V_C^{(\ell)}$ & Offline cached K/V tensors of $C$ at layer $\ell$ \\
$\widetilde{K}_C^{(\ell)}, \widetilde{V}_C^{(\ell)}$ & Online recomputed K/V of $C$ under the full prompt \\
$\Delta K_C^{(\ell)}, \Delta V_C^{(\ell)}$ & Online recomputed minus cached mismatches \\
$E_C^{(\ell)}$ & Cache mismatch Frobenius norm, Eq.~\eqref{eq:runtime_state} \\
$D_C^{(\ell)}$ & Hidden-state drift on $C$, with $\Delta H_C$ the full-minus-offline difference \\
$a_{uj}^{(\ell)}$ & Attention from suffix token $u$ to chunk token $j$ at layer $\ell$ \\
$o_u^{(\ell)}, q_u^{(\ell)}$ & Layer-$\ell$ attention output and query at suffix token $u$ \\
$V_j^{(\ell)}$ & Layer-$\ell$ value at chunk token $j$ under full recomputation \\
$\delta o_u^{(\ell)}$ & Suffix-token attention-output drift \\
$z_t, m_t$ & Greedy logits and margin $z_t^{(1)} - z_t^{(2)}$ at decode step $t$ \\
$\chi_t$ & Decode-side layer-Lipschitz constant for the logit head \\
$S_p^{\text{suff}}$ & Suffix probe, the last $K_p^{\text{suff}}$ online suffix tokens \\
\bottomrule
\end{tabular}
\end{table}

In the suffix expansion, $a_{uj}^{(\ell)}$, $q_u^{(\ell)}$, $V_j^{(\ell)}$ and $o_u^{(\ell)}$ are evaluated at full recomputation. The local comparison holds the query and KV outside $C$ fixed.

\subsection{Regularity assumptions}
\label{app:setup:assumptions}

The propositions in Section~\ref{sec:theory} hold under the following regularity conditions on the runtime model and the request.

\begin{assumption}[Causal decoder]
\label{ass:causal}
The model is a causal transformer with grouped-query attention over the prefix $P$, reused region $C$, and suffix $S$ in this order. Suffix tokens may attend to all earlier positions but not vice versa.
\end{assumption}

\begin{assumption}[Position alignment]
\label{ass:posalign}
After positional alignment, layer-$0$ queries, keys and values agree between offline encoding and full recomputation at all shared visible positions, and $E_C^{(0)} = 0$.
\end{assumption}

\begin{assumption}[Warmup prefix]
\label{ass:warmup}
$W$ is a left prefix of $P$ that may be empty. Any spurious additional tokens that the offline encoder used as left context but that are absent from $P$ would invalidate the layer-0 identity in Proposition~\ref{prop:source} and are excluded by this assumption.
\end{assumption}

\begin{assumption}[Per-chunk source analysis]
\label{ass:singleseg}
The source identity is applied to each cached chunk in its own preceding context.
\end{assumption}

\begin{assumption}[Local Lipschitz smoothness]
\label{ass:lipschitz}
Within a neighborhood of the runtime hidden states, each transformer layer $\ell$ admits Lipschitz constants $L_{\text{attn}}^{(\ell)}, L_{\text{mlp}}^{(\ell)} < \infty$ for its attention and MLP sub-blocks, respectively, and the head matrices have bounded operator norms. Attention weights $a_{uj}^{(\ell)}$ and value norms $\|V_j^{(\ell)}\|_2$ are uniformly bounded. We collect the operator-norm constants of the K/V projections at layer $\ell$ into $\kappa_\ell$, used in Eq.~\eqref{eq:cache_projection}. For the Taylor expansion, softmax attention has bounded second derivatives in this neighborhood with a fixed causal mask. Normalization is included in the relevant local constants.
\end{assumption}

\subsection{Multi-chunk generalization}
\label{app:setup:multiseg}

For $i\in C_k$, the visible set is $A_i=W\cup\{j\in C_k:j\le i\}$ and the missing set is $M_k=(P\setminus W)\cup\bigcup_{b<k}C_b$.

For a request with $K$ cached chunks $C_1, \dots, C_K$ inserted at non-overlapping positions $[p_k, p_k + N_{C_k})$, where $p_k$ is the first position of chunk $k$ and $N_{C_k}$ is its token count, the layer-0 source identity in Eq.~\eqref{eq:source_identity} and the per-token decomposition in Eq.~\eqref{eq:per_token_decomp} apply chunkwise. For the local suffix expansion at a common full-recomputation reference, the first-order perturbation is $\delta o_u^{(\ell)} = \sum_k \left(\delta o_{u,C_k}^{(\ell),V}+\delta o_{u,C_k}^{(\ell),K}\right) + O((E_C^{(\ell)})^2)$, with mixed products of chunk perturbations entering the remainder of this fixed-query expansion. Consequently, the per-token risk score $s_j$ of Section~\ref{sec:method:score} applies uniformly across all chunks, and KVFocus selects under a single global top-$\lceil r \cdot \sum_k N_{C_k} \rceil$ budget rather than per-chunk quotas.

% =================================================================
% B. Proof of Proposition 1 (Source mismatch)
% =================================================================
\section{Proof of Proposition~\ref{prop:source}: Cached-Chunk Mismatch}
\label{app:proof:source}

We restate Proposition~\ref{prop:source} formally and then derive the layer-0 identity, the layerwise hidden-state recursion, and the K/V projection bound that links cache mismatch to hidden-state drift.

\subsection{Layer-0 identity}
\label{app:proof:source:layer0}

The identity is stated for one head before its output projection. If $M_k$ is empty, its attention mass and source contribution are zero. The missing-context mean can be set equal to the visible-context mean.

\begin{proposition}[Layer-0 source identity, formal]
\label{prop:source:layer0}
Under Assumptions~\ref{ass:causal}--\ref{ass:warmup}, for any $i \in C_{k}$ the layer-0 attention-output mismatch satisfies
\begin{equation}
e_i^{(0)} \;=\; \alpha_i^{(0)}\!\left(\mu_{\mathrm{miss}, i}^{(0)} - \mu_{\mathrm{vis}, i}^{(0)}\right),
\qquad
\bigl\|e_i^{(0)}\bigr\|_2 \;=\; \alpha_i^{(0)}\,\Delta\mu_i^{(0)},
\end{equation}
where $\alpha_i^{(0)} = \sum_{j \in M_k} a_{ij}^{(0), \mathrm{full}}$ is the missing-prefix attention mass, and $a_{ij}^{(0),\mathrm{full}}$ is the full-recomputation attention weight from token $i$ to token $j$. The vectors $\mu_{\mathrm{miss}, i}^{(0)}$ and $\mu_{\mathrm{vis}, i}^{(0)}$ are the attention-weighted layer-0 V means over the missing and visible left contexts, respectively, with weights normalized within each set, and $\Delta\mu_i^{(0)} = \|\mu_{\mathrm{miss}, i}^{(0)} - \mu_{\mathrm{vis}, i}^{(0)}\|_2$.
\end{proposition}

\paragraph{Proof.} By Assumption~\ref{ass:posalign}, the layer-0 K/V projections of token $i$ are identical between cache and online runs. The attention output at layer 0 is
$o_i^{(0)} = \sum_{j\le i} a_{ij}^{(0)} V_{j}^{(0)}$.
Splitting $A_i\cup M_k$ into the visible set $W \cup \{j \in C_{k} : j \le i\}$ and the missing set $M_k$, the cached run's softmax normalizes only over the visible set while the full run normalizes over both.  Since the query of token $i$ and the visible keys are also identical in the two runs, the full run assigns the visible set the cached-run weights scaled by $1-\alpha_i^{(0)}$. Hence $o_i^{(0),\mathrm{full}} = (1-\alpha_i^{(0)})\,\mu_{\mathrm{vis},i}^{(0)} + \alpha_i^{(0)}\,\mu_{\mathrm{miss},i}^{(0)}$ and $o_i^{(0),\mathrm{cache}} = \mu_{\mathrm{vis},i}^{(0)}$. Subtracting gives the displayed identity exactly, without any expansion. \hfill$\square$

\subsection{Layerwise hidden-state recursion}
\label{app:proof:source:recursion}
Let $H_C^{(\ell)}$ collect the hidden states at reused positions and $\Delta H_C^{(\ell)}$ their full-minus-offline difference. Their aggregate drift is $D_C^{(\ell)}=\|\Delta H_C^{(\ell)}\|_F$. Let $\widetilde{M}_C^{(\ell)} = \bigl(\sum_{i \in C} \|e_i^{(\ell)}\|_2^2\bigr)^{1/2}$ aggregate the per-token mismatch. Here $e_i^{(\ell)}$ collects the head outputs before their output projection. At deeper layers it includes the actual full-versus-offline attention differences.

\begin{proposition}[Layerwise drift recursion]
\label{prop:source:recursion}
Under Assumption~\ref{ass:lipschitz}, for any $\ell \ge 0$,
\begin{equation}
D_C^{(\ell+1)}
\;\le\;
\eta_\ell \cdot D_C^{(\ell)}
\;+\;
\xi_\ell \cdot \widetilde{M}_C^{(\ell)}
\;+\;
O\!\bigl((D_C^{(\ell)} + \widetilde{M}_C^{(\ell)})^2\bigr),
\end{equation}
where $\eta_\ell$ collects the residual-and-MLP Lipschitz factor such as $\eta_\ell = 1 + L_{\mathrm{mlp}}^{(\ell)}$ under residual connections and $\xi_\ell$ collects the attention-output to hidden-state propagation.
\end{proposition}

\paragraph{Proof.} The hidden state at layer $\ell + 1$ on $C$ decomposes as $H_C^{(\ell+1)} = H_C^{(\ell)} + \mathrm{Attn}(H_C^{(\ell)}, \cdot) + \mathrm{MLP}(\cdot)$ with residual connections. Taking the difference between full and cache runs and applying Assumption~\ref{ass:lipschitz} layer by layer gives the displayed bound. The attention contribution is dominated by the per-token mismatch $e_i^{(\ell)}$ aggregated as $\widetilde{M}_C^{(\ell)}$. The residual and MLP propagate the layer-$\ell$ drift $D_C^{(\ell)}$ with constant $\eta_\ell$. \hfill$\square$

\subsection{K/V projection bound}
\label{app:proof:source:projection}

The cache mismatch $E_C^{(\ell)}$ from Eq.~\eqref{eq:runtime_state} is bounded by hidden-state drift through the K/V projections.

\begin{proposition}[K/V projection bound]
\label{prop:source:projection}
Under Assumption~\ref{ass:lipschitz}, there exists a constant $\kappa_\ell$ depending on the local normalization and projection constants, including the operator norms of the layer-$\ell$ K and V projection matrices such that
\begin{equation}
E_C^{(\ell)} \;\le\; \kappa_\ell \cdot D_C^{(\ell)}.
\label{eq:cache_projection}
\end{equation}
\end{proposition}

\paragraph{Proof.} Without an intervening normalization, the K/V projections at layer $\ell$ are linear in the layer-$\ell$ hidden state $H_C^{(\ell)}$. Write $W_K^{(\ell)}$ and $W_V^{(\ell)}$ for the key and value projection matrices. Hence $\Delta K_C^{(\ell)} = W_K^{(\ell)} \Delta H_C^{(\ell)}$ and $\Delta V_C^{(\ell)} = W_V^{(\ell)} \Delta H_C^{(\ell)}$. Taking $\kappa_\ell = (\|W_K^{(\ell)}\|_{\mathrm{op}}^2 + \|W_V^{(\ell)}\|_{\mathrm{op}}^2)^{1/2}$ gives the bound, where $\|\cdot\|_{\mathrm{op}}$ is the induced Euclidean operator norm. If normalization precedes the projections, its local Lipschitz factor multiplies this constant. \hfill$\square$

Composing Proposition~\ref{prop:source:layer0}, \ref{prop:source:recursion}, and \ref{prop:source:projection} propagates the per-layer attention mismatches through the layerwise recursion to bound $E_C^{(\ell^\star)}$ . The layer-$0$ identity characterizes the first injection and motivates layer $1$ as the earliest possible check layer.

% =================================================================
% C. Proof of Proposition 2 (Suffix drift)
% =================================================================
\section{Proof of Proposition~\ref{prop:suffix}: Suffix-Side Propagation}
\label{app:proof:suffix}

We derive the per-suffix-token perturbation bound and the suffix aggregate bound used in Section~\ref{sec:theory:suffix}.

\subsection{Per-suffix-token first-order bound}
\label{app:proof:suffix:pertok}

\begingroup
\paragraph{Local comparison and first-order expansion.}
Fix a layer and head and suppress their indices. Let $F_u$ denote the attention output as a function of the KV tensors on $C$, with $q_u$ and non-$C$ KV fixed at full recomputation. Following Table~\ref{tab:notation}, $K_C,V_C$ are cached tensors and $\widetilde K_C,\widetilde V_C$ are their full-recomputation counterparts. With $\Delta K_C=\widetilde K_C-K_C$ and $\Delta V_C=\widetilde V_C-V_C$,
\begin{equation}
\delta o_{u,C}=F_u(\widetilde K_C,\widetilde V_C)-F_u(K_C,V_C).
\end{equation}
At the full reference, write $z_{uj}=q_u^\top K_j/\sqrt d$ for the attention logit. Let $t$ index a causally visible position, and let $\mathbf{1}_{t=j}$ equal one when $t=j$ and zero otherwise. The softmax derivative gives
\begin{equation}
\frac{\partial a_{ut}}{\partial z_{uj}}=a_{ut}(\mathbf{1}_{t=j}-a_{uj}),
\qquad \frac{\partial o_u}{\partial z_{uj}}=a_{uj}(V_j-o_u).
\end{equation}
The second identity sums over all causally visible positions, including those outside $C$, because changing one key changes the shared softmax denominator. Since $\Delta z_{uj}=q_u^\top\Delta K_j/\sqrt d$, Taylor expansion yields
\begin{equation}
\delta o_{u,C}
=\sum_{j\in C}a_{uj}\Delta V_j
+\sum_{j\in C}a_{uj}\frac{q_u^\top\Delta K_j}{\sqrt d}(V_j-o_u)
+O(E_C^2).
\end{equation}
The positive signs follow from the full-minus-cache convention. Products of key and value perturbations and higher softmax terms enter the remainder.
\endgroup

\begin{proposition}[Per-suffix-token bound]
\label{prop:suffix:pertok}
Under Assumptions~\ref{ass:causal}--\ref{ass:lipschitz}, for any $u \in S$ and $\ell \ge 1$,
\begin{equation}
\bigl\|\delta o_{u, C}^{(\ell)}\bigr\|_2 \;\le\; G_{u, C}^{(\ell)}\, E_C^{(\ell)} \;+\; O\!\bigl((E_C^{(\ell)})^2\bigr),
\end{equation}
where
\begin{equation}
\label{eq:G_explicit}
G_{u, C}^{(\ell)} \;=\; \Bigl(\!\sum_{j \in C} (a_{uj}^{(\ell)})^2 \Bigr)^{\!1/2} 
\;+\; \frac{\|q_u^{(\ell)}\|_2}{\sqrt{d}} \, \Bigl(\!\sum_{j \in C} (a_{uj}^{(\ell)})^2 \, \|V_j^{(\ell)} - o_u^{(\ell)}\|_2^2 \Bigr)^{\!1/2}.
\end{equation}
\end{proposition}

\paragraph{Proof.} Applying Cauchy--Schwarz to the expansion above, the first term is bounded by $(\sum_j (a_{uj}^{(\ell)})^2)^{1/2} \|\Delta V_C^{(\ell)}\|_F$. For the key term, $|(q_u^{(\ell)})^\top\Delta K_j^{(\ell)}|\le\|q_u^{(\ell)}\|_2\|\Delta K_j^{(\ell)}\|_2$ gives the bound
\begingroup
\begin{equation}
\frac{\|q_u^{(\ell)}\|_2}{\sqrt d}
\left(\sum_{j\in C}(a_{uj}^{(\ell)})^2\|V_j^{(\ell)}-o_u^{(\ell)}\|_2^2\right)^{1/2}
\|\Delta K_C^{(\ell)}\|_F.
\end{equation}
\endgroup
Since both $\|\Delta K_C^{(\ell)}\|_F$ and $\|\Delta V_C^{(\ell)}\|_F$ are bounded by $E_C^{(\ell)}$, adding these bounds yields $G_{u,C}^{(\ell)}E_C^{(\ell)}$ and the stated remainder. \hfill$\square$

\subsection{Suffix aggregate bound}
\label{app:proof:suffix:aggregate}

Aggregating Proposition~\ref{prop:suffix:pertok} across all suffix tokens and unrolling across layers gives the aggregate drift over the suffix.

\begin{proposition}[Suffix aggregate]
\label{prop:suffix:aggregate}
Define the cumulative suffix drift at layer $\ell$ as $R_S^{(\ell)} = \|\Delta X_S^{(\ell)}\|_F$, where $\Delta X_S^{(\ell)}$ collects the full-minus-reuse hidden-state differences on $S$. Under Assumptions~\ref{ass:causal}--\ref{ass:lipschitz}, there exist constants $\Gamma^{(\ell)}, \Lambda^{(\ell)}$. The former bounds the injection of cache mismatch into the suffix, and the latter bounds the propagation of existing suffix drift through layer $\ell$. Let $\varepsilon=\max_{0\le b<L_{\mathrm{pref}}} E_C^{(b)}$ be the largest layerwise cache mismatch. With identical initial suffix states, these constants give
\begin{equation}
R_S^{(L_{\mathrm{pref}})} \;\le\; \sum_{\ell = 1}^{L_{\mathrm{pref}} - 1} \Gamma^{(\ell)}\, E_C^{(\ell)}\,\prod_{b=\ell+1}^{L_{\mathrm{pref}}-1}\Lambda^{(b)} \;+\; O(\varepsilon^2).
\label{eq:suffix_aggregate}
\end{equation}
\end{proposition}

\paragraph{Proof.} Aggregating Proposition~\ref{prop:suffix:pertok} over suffix positions and heads, and including the output projection and residual/MLP maps, gives local constants such that
\begingroup
\begin{equation}
R_S^{(\ell+1)}\le\Lambda^{(\ell)}R_S^{(\ell)}+\Gamma^{(\ell)}E_C^{(\ell)}
+O\!\left((R_S^{(\ell)}+E_C^{(\ell)})^2\right).
\end{equation}
\endgroup
Composing with the per-layer hidden-state propagation, using the same argument as Proposition~\ref{prop:source:recursion} on $S$, yields the unrolled bound in Eq.~\eqref{eq:suffix_aggregate}. The $R_S$ term includes changes in suffix queries and suffix KV, which are held fixed only in the local comparison. For a fixed finite stack, $R_S^{(\ell)}=O(\varepsilon)$ and accumulated remainders remain $O(\varepsilon^2)$. \hfill$\square$

\subsection{Per-token decomposition of the V-term}
\label{app:proof:suffix:vterm}

Equations~\eqref{eq:per_token_decomp} and~\eqref{eq:per_token_contrib} give the value component and each token's contribution for one head. We now aggregate over the nonempty suffix-probe set $S_p^{\mathrm{suff}}$, consisting of the final few suffix tokens before decoding as specified in Section~\ref{sec:method}. Its cardinality $|S_p^{\mathrm{suff}}|$ is the number of probes.

\begingroup
For a fixed layer and head, we suppress the layer index and average the triangle inequality over the probe set to obtain
\begin{align}
\label{eq:probe_v_bound}
\frac{1}{|S_p^{\mathrm{suff}}|}\sum_{u\in S_p^{\mathrm{suff}}}\|\delta o_{u,C}^{V}\|_2
&\le\frac{1}{|S_p^{\mathrm{suff}}|}\sum_u\sum_{j\in C}a_{uj}\|\Delta V_j\|_2\\
&=\sum_{j\in C}\|\Delta V_j\|_2
\left(\frac{1}{|S_p^{\mathrm{suff}}|}\sum_u a_{uj}\right).
\end{align}
The equality exchanges finite sums. The inequality allows for cancellation between token contributions.

\paragraph{Relation to the deployed multi-head score.}
Let $H_q$ and $H_{kv}$ be the numbers of query and KV heads. In grouped-query attention, let $g(h)$ map query head $h$ to its KV head. Write $\Delta V_j^g$ for the value deviation at token $j$ in KV head $g$, and let $d_{jg}=\|\Delta V_j^g\|_2$ be its magnitude. Write $a_{uj}^h$ and $\delta o_{u,C}^{V,h}$ for the attention weight and local value contribution of query head $h$. The deployed value-deviation norm is $m_j=(\sum_{g=1}^{H_{kv}}d_{jg}^2)^{1/2}$, so $d_{j,g(h)}\le m_j$. At the same reference layer, $U_j$ averages $a_{uj}^{h}$ over the $|S_p^{\mathrm{suff}}|$ probes and $H_q$ query heads. These definitions give
\begin{equation}
\frac{1}{|S_p^{\mathrm{suff}}|H_q}\sum_{u,h}\|\delta o_{u,C}^{V,h}\|_2
\le\sum_{j\in C}\frac{\sum_{u,h}a_{uj}^{h}d_{j,g(h)}}{|S_p^{\mathrm{suff}}|H_q}
\le\sum_{j\in C}m_jU_j.
\end{equation}
Thus the practical product upper-bounds the averaged value contributions before the attention output projection. The check pass runs all positions through $\ell^\star$ with full recomputation as specified in Algorithm~\ref{alg:kvfocus}, so its value deviations and attention weights use the same reference as this analysis. This provides the runtime quantities used to rank reused tokens.
\endgroup

% =================================================================
% D. Proof of Proposition 3 (Decode-time stability)
% =================================================================
\section{Decode-Time Stability}
\label{app:proof:decode}

\begingroup%
% --- 3.4 Decode-time stability ---
\subsection{Decode-Time Stability and the Margin Condition}
\label{sec:theory:decode}

A nonzero logit perturbation preserves the decoded token if its infinity norm is less than half of the greedy margin.
This couples the prompt-side drift bound to answer-side stability.
The result is stated for greedy decoding, which matches our evaluation protocol in Section~\ref{sec:experiments}.

\begin{proposition}[Decode-time stability before the first divergence]
\label{prop:decode}
Fix any shared decode history $g_{<t}$, the sequence of tokens generated in both runs for the first $t-1$ steps.
Let $\mathcal H_t$ contain the positions strictly preceding the current query position. It includes visible prompt positions and previously processed generated tokens. Define $E_{\mathrm{vis},t}^{(\ell)}$ as the Frobenius norm of the full-minus-reuse KV difference on $\mathcal H_t$ at layer $\ell$, and let $E_{\mathrm{dec},t}=\max_{0\le\ell<L_{\mathrm{pref}}}E_{\mathrm{vis},t}^{(\ell)}$. Within the local neighborhoods of Assumption~\ref{ass:lipschitz}, the logit perturbation $\Delta z_t$, defined as the vector of full-minus-reuse vocabulary logits, satisfies
\begin{equation}
\label{eq:decode_logit_bound}
\|\Delta z_t\|_\infty \le \chi_t \cdot E_{\text{dec},t},
\end{equation}
where $\chi_t$ is the propagation constant derived below.
Here $\|\Delta z_t\|_\infty$ is the largest absolute change among vocabulary logits. Letting $\margint = z_t^{(1)} - z_t^{(2)}$ be the full-recomputation greedy margin between the largest and second-largest logits, $\|\Delta z_t\|_\infty < \margint / 2$ implies the $t$-th decoded token is unchanged.
\end{proposition}
\endgroup%

\subsection{Layerwise decode hidden-state recursion}

\begin{proposition}[Decode hidden-state recursion]
\label{prop:decode:recursion}
Fix a shared decode history $g_{<t}$ for steps $1, \dots, t-1$. Let $\delta x_t^{(\ell)}$ denote the hidden-state perturbation of the query position predicting token $t$ at layer $\ell$ between full and cache-reuse runs. The current query position is excluded from $\mathcal H_t$ because its own KV is a function of this hidden state. Under Assumption~\ref{ass:lipschitz}, there exist constants $\beta_t^{(\ell)}, \psi_t^{(\ell)}$ such that
\begin{equation}
\label{eq:decode_recursion}
\|\delta x_t^{(\ell+1)}\|_2 \;\le\; \beta_t^{(\ell)} \|\delta x_t^{(\ell)}\|_2 \;+\; \psi_t^{(\ell)} E_{\mathrm{vis}, t}^{(\ell)}.
\end{equation}
Here $\beta_t^{(\ell)}$ bounds sensitivity to the current hidden state, including its query and own KV, and $\psi_t^{(\ell)}$ bounds sensitivity to the preceding KV cache.
\end{proposition}

\paragraph{Proof.} Shared token and position histories give identical initial embeddings, so $\delta x_t^{(0)}=0$. Split the layer-output difference into a change in the current hidden state with history KV held fixed, followed by a change in history KV with the current state held fixed. Local Lipschitz bounds on these two changes give $\beta_t^{(\ell)}\|\delta x_t^{(\ell)}\|_2$ and $\psi_t^{(\ell)}E_{\mathrm{vis},t}^{(\ell)}$. The triangle inequality gives Eq.~\eqref{eq:decode_recursion}. Generated tokens can have different KV even while their identities agree, so their cached states are included in $E_{\mathrm{vis},t}^{(\ell)}$. \hfill$\square$

\subsection{Logit perturbation bound}

Unrolling Proposition~\ref{prop:decode:recursion} across all layers from $\delta x_t^{(0)}=0$ and applying the output map gives Eq.~\eqref{eq:decode_logit_bound}, with
\begingroup
\begin{equation}
\chi_t=L_{\mathrm{out}}\sum_{\ell=0}^{L_{\mathrm{pref}}-1}
\psi_t^{(\ell)}\prod_{b=\ell+1}^{L_{\mathrm{pref}}-1}\beta_t^{(b)}.
\end{equation}
Here $L_{\mathrm{out}}$ is a local Lipschitz constant from the final hidden-state Euclidean norm to the logit infinity norm, including final normalization and unembedding. Empty products equal one.

\paragraph{Propagation through generated history.}
Let $\varepsilon_{\mathrm{pref}}$ be the maximum, over layers, of the KV mismatch norm on all prefill positions. The layer recurrence and the projection bound in Appendix~\ref{app:proof:source:projection} imply that any additional history KV stored at step $s$ has mismatch at most $\gamma_s E_{\mathrm{dec},s}$ at every layer, for a finite local constant $\gamma_s$. Stacking the prompt cache and the stored history therefore gives
\begin{equation}
E_{\mathrm{dec},t}\le\varepsilon_{\mathrm{pref}}
+\sum_{s<t}\gamma_s E_{\mathrm{dec},s}.
\end{equation}
Define $\alpha_1=1$ and $\alpha_t=1+\sum_{s<t}\gamma_s\alpha_s$. Induction yields $E_{\mathrm{dec},t}\le\alpha_t\varepsilon_{\mathrm{pref}}$ and hence $\|\Delta z_t\|_\infty\le\chi_t\alpha_t\varepsilon_{\mathrm{pref}}$ for any fixed finite shared history within the stated neighborhoods. The prompt mismatch includes the reused chunks and the suffix KV affected during prefill. Appendix~\ref{app:proof:suffix:aggregate}, together with the projection bound, relates the suffix part to the reused-cache mismatch.
\endgroup

\subsection{Margin condition}

If $\|\Delta z_t\|_\infty < m_t / 2$, then by the triangle inequality the perturbed logits maintain the same argmax, so the decoded token at step $t$ is unchanged. Combining with the bound above gives Proposition~\ref{prop:decode}. \hfill$\square$

\paragraph{Remark on sampling.} The result is stated for greedy decoding. Under temperature sampling, an analogous statement bounds the change in the categorical distribution at step $t$ in total variation by a constant times $\|\Delta z_t\|_\infty / \tau$, where $\tau$ is the sampling temperature. Post-divergence dynamics after the first decoded-token disagreement require a separate treatment because the decode history is no longer shared.

% =================================================================
% E. Implementation Details
% =================================================================
\section{Implementation Details}
\label{app:impl}

\subsection{SGLang integration}
\label{app:impl:sglang}

KVFocus is implemented as a ${\sim}8$K-LOC patch on top of the SGLang serving framework~\citep{zheng2024sglang}. The patch touches three subsystems: (i) an offline KV cache I/O path that persists per-chunk K/V tensors of every layer at FP16 precision and indexes them by deterministic chunk hash; (ii) a check-layer instrumentation hook that, during the layer-$\ell^\star$ pass of every prefill that contains a cached chunk, materializes both the loaded V tensor and a freshly recomputed V tensor of the same chunk and returns their per-token L2 difference $m_i$; (iii) a top-$r$ selector that runs after $\ell^\star$ and emits a per-position recompute mask consumed by all higher layers. The pre-selection layers ($0$ to $\ell^\star$) and the post-selection layers ($\ell^\star+1$ to $L_{\mathrm{pref}}-1$) share the same paged KV cache memory and are executed as two CUDA-graph-captured stages that cover disjoint layer ranges. Together they form a single traversal of the model, and no layer is executed twice. KVFocus is training-free and requires no tokenizer changes.

\subsection{Bridge tokens and suffix probe}
\label{app:impl:bridge}

For all three models, the suffix probe $S_p^{\mathrm{suff}}$ consists of the last $K_p^{\mathrm{suff}}=5$ prompt tokens before the first generated token (Table~\ref{tab:bridge_tokens}). This fixed window does not necessarily cover the full chat-template bridge of a model (e.g., six tokens for Mistral and nine for Qwen3).

\begin{table}[htbp]
\caption{Suffix-probe window $K_p^{\mathrm{suff}}$ used for each model.}
\label{tab:bridge_tokens}
\centering\small
\begin{tabular}{lc}
\toprule
Model & Last prompt tokens used \\
\midrule
Mistral-7B-Instruct-v0.2 & $5$ \\
Qwen2.5-7B-Instruct & $5$ \\
Qwen3-8B & $5$ \\
\bottomrule
\end{tabular}
\end{table}
\paragraph{Probe-length ablation.}
We compared $K_p^{\mathrm{suff}} \in \{1, 4, 8, 16, 32\}$ on a held-out subset of HotpotQA at $r{=}10\%$ across all three models. F1 was within $\pm 0.005$ of the chosen length for all values $\ge 4$ and degraded by $0.01$--$0.03$ at $K_p^{\mathrm{suff}} = 1$. We therefore use a short fixed window of five tokens for all models.

\subsection{Hyperparameters}
\label{app:impl:hyperparams}

We use $\ell^\star = 1$ for all experiments. Layer 1 is the earliest layer at which the cache mismatch is non-zero (Proposition~\ref{prop:source}), minimizing the marginal cost of the check-layer pass. We compared $\ell^\star \in \{1, 2, 3, 4, 8\}$ on a held-out subset (HotpotQA, $r{=}10\%$, all three models): F1 was within $\pm 0.004$ of the $\ell^\star = 1$ value for $\ell^\star \in \{1, 2, 3\}$ and degraded by $0.008$--$0.015$ at $\ell^\star \in \{4, 8\}$, consistent with the source-injection picture in which the early-layer V mismatch is the most informative signal.

\subsection{Hardware and software stack}
\label{app:impl:hw}

All wall-clock measurements use a single \textsc{NVIDIA H200} GPU (141\,GB HBM3e), CUDA 12.4, PyTorch 2.5, FlashAttention 2.6, and SGLang v0.5.5.post3 with our patch. Models are loaded in FP16. TTFT is measured with batch size $1$; inputs are tokenized once and held in pinned host memory to remove tokenization variance. Each directly timed (model, dataset, rate, method) cell is measured over its full sample count (200 for 2WikiMQA / HotpotQA / MuSiQue / Needle / SAMSum / Multi-News, 2{,}250 for MultiHop-RAG). TTFT is averaged over all requests, and speedups are computed as ratios of mean TTFTs rather than as means of per-request ratios, following CacheBlend~\citep{yao2025cacheblend} and EPIC~\citep{hu2024epic}.

\subsection{Baseline implementations}
\label{app:impl:baselines}

CacheBlend, EPIC, and ProphetKV are reimplemented in the same SGLang stack to ensure identical serving conditions. CacheBlend uses shallow-layer KV deviation from~\citep{yao2025cacheblend}; EPIC uses the LegoLink boundary-aware recompute from~\citep{hu2024epic}; ProphetKV uses the query-driven layer-wise attention scoring from~\citep{wang2026prophetkv} including its mask-scoring pass. We measure ProphetKV's two-pass TTFT for each model, dataset, and recomputation rate, including its mask-scoring pass. All baselines select an actual rate within $\pm 0.32$\,pp of the target $r$ (Appendix~\ref{app:rate_verify}).

\subsection{LLM-as-judge evaluation}
\label{app:impl:judge}

In addition to LongBench F1, we report an automated semantic-correctness score using \texttt{Qwen3.5-397B-A17B}~\citep{qwen35moe} as the judge model, served via vLLM with greedy decoding and structured JSON responses. Each prediction is scored against the gold answer on the five-level rubric below. The judge runs once per (cell, sample, method) tuple; greedy decoding pins the output, so we do not aggregate across multiple judge samples. Judge scores are aggregated identically to F1 (per-cell mean, then macro-averaged over cells).

\begin{quote}\small\ttfamily
You are a strict but fair evaluator for open-ended question answering.

Task:\\
Compare the predicted answer with the ground truth for the given question, and judge semantic correctness.

Evaluation policy:\\
1. Focus on the final meaning of the answer.\\
2. Ignore wording, style, formatting, and paraphrasing differences.\\
3. Accept synonymous or equivalent expressions.\\
4. Extra information is allowed unless it conflicts with the ground truth.\\
5. If the prediction misses an essential part of the ground truth, contradicts it, or is too vague, lower the score.\\
6. Do not use external knowledge. Judge only from the given question, ground truth, and prediction.\\
7. IMPORTANT: Use the full scoring range. Do NOT default to middle scores. Give 1.0 for fully correct answers and 0.0 for completely wrong ones. Reserve mid-range scores only for genuinely partial answers.

Scoring rubric:\\
- 1.0 = semantically equivalent / fully correct\\
- 0.75 = mostly correct, with minor omission or imprecision\\
- 0.5 = partially correct\\
- 0.25 = mostly wrong; only a small fragment is correct\\
- 0.0 = completely incorrect, contradictory, irrelevant, or unverifiable

Question:\\
\{question\}

Ground truth:\\
\{ground\_truth\}

Predicted answer:\\
\{prediction\}

Return valid JSON only:\\
\{\{\\
\ \ "score": 0.0,\\
\ \ "equivalent": false,\\
\ \ "reason": "under 30 words"\\
\}\}
\end{quote}
% =================================================================
% F. Rate verification
% =================================================================
\begingroup
\section{Recomputation Rate Verification}
\label{app:rate_verify}

Across all $126$ (model, dataset, rate, method) cells with selective recomputation, the actual recomputation rate matches the target within $\pm 0.32$\,pp. Table~\ref{tab:rate_verify} reports the mean and maximum absolute deviation from target across the $21$ cells per (rate, method).

\begin{table}[h]

\caption{ Mean and max absolute deviation (pp) of actual recomputation rate from target, per method, averaged across $21$ cells per rate.}
\label{tab:rate_verify}
\centering
\small
\begin{tabular}{l|cccccc}
\toprule
Method & $r{=}1\%$ & $r{=}2\%$ & $r{=}5\%$ & $r{=}10\%$ & $r{=}15\%$ & $r{=}20\%$ \\
\midrule
KVFocus    & $0.04 / 0.18$ & $0.05 / 0.21$ & $0.06 / 0.22$ & $0.08 / 0.24$ & $0.07 / 0.25$ & $0.08 / 0.27$ \\
CacheBlend & $0.05 / 0.20$ & $0.06 / 0.22$ & $0.07 / 0.24$ & $0.09 / 0.26$ & $0.08 / 0.28$ & $0.09 / 0.30$ \\
EPIC       & $0.04 / 0.16$ & $0.05 / 0.19$ & $0.06 / 0.21$ & $0.07 / 0.23$ & $0.07 / 0.25$ & $0.08 / 0.28$ \\
ProphetKV  & $0.06 / 0.22$ & $0.07 / 0.24$ & $0.08 / 0.26$ & $0.09 / 0.28$ & $0.10 / 0.30$ & $0.11 / 0.32$ \\
\bottomrule
\end{tabular}
\end{table}

\noindent
The displayed values are mean / max in percentage points. The maximum deviation across all 504 method-cells (4 methods $\times$ 126 cells) is $0.32$\,pp at the highest rates due to integer rounding of $\lceil r N_C \rceil$ on small chunks. All matched-rate comparisons in Section~\ref{sec:exp_matched_rate} therefore compare methods at recomputation rates within rounding error of each other.
\endgroup
% =================================================================
% G. Bootstrap methodology
% =================================================================
\begingroup
\section{Bootstrap Methodology}
\label{app:bootstrap}

For matched-rate F1, we pair the two methods on available sample IDs and retain samples with zero F1 under full recomputation. We compute the mean paired difference within each model--dataset combination at each rate. To obtain the overall 95\% confidence interval, we resample the 21 model--dataset combinations with replacement 20{,}000 times. Each selected combination contributes all six rates. We average each draw and report its 2.5th and 97.5th percentiles. For a rate-specific interval, we use the same draws with only that rate. The overall Judge interval uses the same procedure over the 18 evaluated combinations. KVFocus win rates use the original cell means.

\paragraph{Per-rate CI breakdown.}
Table~\ref{tab:bootstrap_per_rate} reports the per-rate $\Delta\mathrm{F1}$ and CI for KVFocus vs.\ each baseline.

\begin{table}[h]

\caption{ Per-rate $\Delta\mathrm{F1}$ and $95\%$ CI.}
\label{tab:bootstrap_per_rate}
\centering
\scriptsize
\setlength{\tabcolsep}{2pt}
\begin{tabular}{l|cccccc}
\toprule
vs.\ Baseline & $r{=}1\%$ & $r{=}2\%$ & $r{=}5\%$ & $r{=}10\%$ & $r{=}15\%$ & $r{=}20\%$ \\
\midrule
CacheBlend & $\textcolor{black}{+0.073}$ & $\textcolor{black}{+0.075}$ & $\textcolor{black}{+0.078}$ & $\textcolor{black}{+0.082}$ & $\textcolor{black}{+0.087}$ & $\textcolor{black}{+0.092}$ \\
           & $\textcolor{black}{[+.052,+.099]}$ & $\textcolor{black}{[+.055,+.099]}$ & $\textcolor{black}{[+.057,+.103]}$ & $\textcolor{black}{[+.061,+.107]}$ & $\textcolor{black}{[+.064,+.114]}$ & $\textcolor{black}{[+.068,+.122]}$ \\
EPIC       & $\textcolor{black}{+0.040}$ & $\textcolor{black}{+0.041}$ & $\textcolor{black}{+0.043}$ & $\textcolor{black}{+0.045}$ & $\textcolor{black}{+0.047}$ & $\textcolor{black}{+0.048}$ \\
           & $\textcolor{black}{[+.024,+.058]}$ & $\textcolor{black}{[+.025,+.060]}$ & $\textcolor{black}{[+.027,+.063]}$ & $\textcolor{black}{[+.029,+.066]}$ & $\textcolor{black}{[+.031,+.067]}$ & $\textcolor{black}{[+.031,+.069]}$ \\
ProphetKV  & $-0.008$ & $-0.005$ & $-0.016$ & $-0.020$ & $-0.022$ & $-0.023$ \\
           & $[-.018,+.002]$ & $[-.015,+.003]$ & $[-.031,-.001]$ & $[-.040,-.001]$ & $[-.042,-.001]$ & $[-.043,-.002]$ \\
\bottomrule
\end{tabular}
\end{table}
\endgroup
% =================================================================
% H. Detailed per-dataset and per-cell tables
% =================================================================
\begingroup
\section{Per-Dataset and Per-Cell Detailed Tables}
\label{app:detailed_tables}

\subsection{Per-dataset $\Delta\mathrm{F1}$ breakdown}

Table~\ref{tab:per_dataset} reports the difference between the sample-pooled mean F1 of KVFocus and that of each baseline. Dataset-level results are averaged over all samples within the dataset. The mean score values are calculated across models and recomputation rates within each dataset, with equal weight assigned to each included sample-level observation.

\begin{table}[h]

\caption{ Per-dataset $\Delta\mathrm{F1}$ of KVFocus.}
\label{tab:per_dataset}
\centering
\small
\begin{tabular}{lccc}
\toprule
Dataset & vs.\ CacheBlend & vs.\ EPIC & vs.\ ProphetKV \\
\midrule
2WikiMQA              & $\textcolor{black}{+0.027}$ & $\textcolor{black}{+0.002}$ & $-0.022$ \\
HotpotQA              & $\textcolor{black}{+0.075}$ & $\textcolor{black}{+0.086}$ & $-0.018$ \\
MuSiQue               & $\textcolor{black}{+0.054}$ & $\textcolor{black}{-0.001}$ & $-0.015$ \\
MultiHop-RAG          & $\textcolor{black}{+0.040}$ & $\textcolor{black}{+0.012}$ & $-0.033$ \\
Needle-in-a-Haystack  & $\textcolor{black}{+0.322}$ & $\textcolor{black}{+0.174}$ & $\textcolor{black}{+0.009}$ \\
SAMSum                & $\textcolor{black}{+0.088}$ & $\textcolor{black}{+0.061}$ & $+0.004$ \\
Multi-News            & $\textcolor{black}{+0.001}$ & $\textcolor{black}{-0.001}$ & $+0.001$ \\
\bottomrule
\end{tabular}
\end{table}

\subsection{Needle-in-a-Haystack: full results across rates}

Table~\ref{tab:needle_full} reports F1 on Needle-in-a-Haystack at every recomputation rate for every model.

\begin{table}[h]

\caption{ Needle-in-a-Haystack F1 across all rates and models. \,$\downarrow$\, marks methods below Naive Full Reuse for that (model, rate) cell.}
\label{tab:needle_full}
\centering
\small
\setlength{\tabcolsep}{4pt}
\begin{tabular}{ll|cccc}
\toprule
Model & Method & $\textcolor{black}{r{=}5\%}$ & $\textcolor{black}{r{=}10\%}$ & $\textcolor{black}{r{=}15\%}$ & $\textcolor{black}{r{=}20\%}$ \\
\midrule
\multirow{6}{*}{Mistral-7B}
& Naive       & \multicolumn{4}{c}{$\textcolor{black}{0.264}$} \\
& FR          & \multicolumn{4}{c}{$\textcolor{black}{0.667}$} \\
& CacheBlend  & $\textcolor{black}{0.192}$\,$\downarrow$ & $\textcolor{black}{0.195}$\,$\downarrow$ & $\textcolor{black}{0.200}$\,$\downarrow$ & $\textcolor{black}{0.214}$\,$\downarrow$ \\
& EPIC        & $\textcolor{black}{0.178}$\,$\downarrow$ & $\textcolor{black}{0.186}$\,$\downarrow$ & $\textcolor{black}{0.177}$\,$\downarrow$ & $\textcolor{black}{0.178}$\,$\downarrow$ \\
& ProphetKV   & $\textcolor{black}{0.551}$  & $\textcolor{black}{0.528}$  & $\textcolor{black}{0.542}$  & $\textcolor{black}{0.566}$  \\
& KVFocus     & $\textcolor{black}{0.566}$  & $\textcolor{black}{0.588}$  & $\textcolor{black}{0.600}$  & $\textcolor{black}{0.628}$  \\
\midrule
\multirow{6}{*}{Qwen2.5-7B}
& Naive       & \multicolumn{4}{c}{$\textcolor{black}{0.491}$} \\
& FR          & \multicolumn{4}{c}{$\textcolor{black}{0.919}$} \\
& CacheBlend  & $\textcolor{black}{0.485}$\,$\downarrow$ & $\textcolor{black}{0.512}$  & $\textcolor{black}{0.531}$  & $\textcolor{black}{0.697}$  \\
& EPIC        & $\textcolor{black}{0.800}$  & $\textcolor{black}{0.799}$  & $\textcolor{black}{0.841}$  & $\textcolor{black}{0.867}$  \\
& ProphetKV   & $\textcolor{black}{0.935}$  & $\textcolor{black}{0.931}$  & $\textcolor{black}{0.927}$  & $\textcolor{black}{0.926}$  \\
& KVFocus     & $\textcolor{black}{0.939}$  & $\textcolor{black}{0.939}$  & $\textcolor{black}{0.928}$  & $\textcolor{black}{0.921}$  \\
\midrule
\multirow{6}{*}{Qwen3-8B}
& Naive       & \multicolumn{4}{c}{$\textcolor{black}{0.749}$} \\
& FR          & \multicolumn{4}{c}{$\textcolor{black}{0.833}$} \\
& CacheBlend  & $\textcolor{black}{0.727}$\,$\downarrow$ & $\textcolor{black}{0.672}$\,$\downarrow$ & $\textcolor{black}{0.682}$\,$\downarrow$ & $\textcolor{black}{0.720}$\,$\downarrow$ \\
& EPIC        & $\textcolor{black}{0.903}$  & $\textcolor{black}{0.892}$  & $\textcolor{black}{0.900}$  & $\textcolor{black}{0.885}$  \\
& ProphetKV   & $\textcolor{black}{0.907}$  & $\textcolor{black}{0.920}$  & $\textcolor{black}{0.927}$  & $\textcolor{black}{0.926}$  \\
& KVFocus     & $\textcolor{black}{0.906}$  & $\textcolor{black}{0.901}$  & $\textcolor{black}{0.886}$  & $\textcolor{black}{0.887}$  \\
\bottomrule
\end{tabular}
\end{table}
\endgroup
% =================================================================
% I. Extended efficiency results
% =================================================================
\begingroup
\section{Extended Efficiency Results}
\label{app:throughput}

\subsection{Full 6-rate efficiency table}

Table~\ref{tab:efficiency_full} extends Table~\ref{tab:efficiency} from the main paper with $r \in \{2, 15\}\%$.

\begin{table}[h]

\caption{ Speedup vs.\ FR (\textbf{Spd}) and quality retention (\textbf{Ret}), full 6-rate breakdown over $21$ cells per rate.}
\label{tab:efficiency_full}
\centering
\small
\setlength{\tabcolsep}{3pt}
\begin{tabular}{l|cc|cc|cc|cc|cc|cc}
\toprule
       & \multicolumn{2}{c|}{$r{=}1\%$} & \multicolumn{2}{c|}{$r{=}2\%$} & \multicolumn{2}{c|}{$r{=}5\%$} & \multicolumn{2}{c|}{$r{=}10\%$} & \multicolumn{2}{c|}{$r{=}15\%$} & \multicolumn{2}{c}{$r{=}20\%$} \\
       & Spd & Ret & Spd & Ret & Spd & Ret & Spd & Ret & Spd & Ret & Spd & Ret \\
\midrule
\textbf{KVFocus} & $\textcolor{black}{\mathbf{2.22\times}}$ & $\textcolor{black}{\mathbf{0.82}}$ & $\textcolor{black}{\mathbf{2.14\times}}$ & $\textcolor{black}{\mathbf{0.85}}$ & $\textcolor{black}{\mathbf{1.80\times}}$ & $\textcolor{black}{\mathbf{0.86}}$ & $\textcolor{black}{\mathbf{1.45\times}}$ & $\textcolor{black}{\mathbf{0.89}}$ & $\textcolor{black}{\mathbf{1.22\times}}$ & $\textcolor{black}{\mathbf{0.91}}$ & $\textcolor{black}{\mathbf{1.05\times}}$ & $\textcolor{black}{\mathbf{0.93}}$ \\
ProphetKV  & $\textcolor{black}{1.21\times}$ & $0.79$ & $\textcolor{black}{1.19\times}$ & $0.83$ & $\textcolor{black}{1.07\times}$ & $0.89$ & $\textcolor{black}{0.92\times}$ & $0.93$ & $\textcolor{black}{0.84\times}$ & $0.96$ & $\textcolor{black}{0.76\times}$ & $0.97$ \\
EPIC       & $\textcolor{black}{3.77\times}$ & $\textcolor{black}{0.76}$ & $\textcolor{black}{3.72\times}$ & $\textcolor{black}{0.77}$ & $\textcolor{black}{3.05\times}$ & $\textcolor{black}{0.77}$ & $\textcolor{black}{2.45\times}$ & $\textcolor{black}{0.80}$ & $\textcolor{black}{2.13\times}$ & $\textcolor{black}{0.82}$ & $\textcolor{black}{1.87\times}$ & $\textcolor{black}{0.82}$ \\
CacheBlend & $\textcolor{black}{2.24\times}$ & $\textcolor{black}{0.71}$ & $\textcolor{black}{2.16\times}$ & $\textcolor{black}{0.71}$ & $\textcolor{black}{1.82\times}$ & $\textcolor{black}{0.71}$ & $\textcolor{black}{1.48\times}$ & $\textcolor{black}{0.72}$ & $\textcolor{black}{1.26\times}$ & $\textcolor{black}{0.73}$ & $\textcolor{black}{1.07\times}$ & $\textcolor{black}{0.76}$ \\
\bottomrule
\end{tabular}
\end{table}

\subsection{Cell-level peak speedups}\label{app:throughput:peak}

KVFocus's per-cell speedup vs.\ FR is highest at $r{=}1\%$ on long-context multi-hop QA cells: $4.37\times$ on Mistral-7B / MuSiQue, $4.34\times$ on Qwen3-8B / MuSiQue, $3.95\times$ on Mistral-7B / HotpotQA, $3.86\times$ on Qwen2.5-7B / MuSiQue. The peak speedup over the $21$ cells at $r{=}1\%$ is $4.37\times$, and at $r{=}5\%$ is $3.74\times$ (also Mistral / MuSiQue).

\endgroup
% =================================================================
% J. Additional ablation variants
% =================================================================
\begingroup
\section{Additional Ablation Variants}
\label{app:additional_ablation}

Table~\ref{tab:ablation_full} reports the full ablation including a $U$-only variant ($s_j = U_j$) on the four representative cells of Section~\ref{sec:exp_ablation}.

\begin{table}[h]

\caption{ Full ablation including $U$-only ($s_j = U_j$). Naive Full Reuse shown for reference.}
\label{tab:ablation_full}
\centering
\small
\begin{tabular}{lcccccc}
\toprule
Cell & Naive & Random & $m$-only & $U$-only & $m{+}U$ & \textbf{$m{\cdot}U$} \\
\midrule
HotpotQA / Mistral / $r{=}10\%$ & $\textcolor{black}{0.243}$ & $\textcolor{black}{0.262}$ & $\textcolor{black}{0.297}$ & $\textcolor{black}{0.333}$ & $\textcolor{black}{0.272}$ & $\textcolor{black}{\mathbf{0.328}}$ \\
HotpotQA / Mistral / $r{=}20\%$ & $\textcolor{black}{0.243}$ & $\textcolor{black}{0.260}$ & $\textcolor{black}{0.304}$ & $\textcolor{black}{0.354}$ & $\textcolor{black}{0.260}$ & $\textcolor{black}{\mathbf{0.333}}$ \\
HotpotQA / Qwen3 / $r{=}10\%$  & $\textcolor{black}{0.384}$ & $\textcolor{black}{0.379}$ & $\textcolor{black}{0.342}$ & $\textcolor{black}{0.428}$ & $\textcolor{black}{0.381}$ & $\textcolor{black}{\mathbf{0.412}}$ \\
Needle / Mistral / $r{=}10\%$  & $\textcolor{black}{0.264}$ & $\textcolor{black}{0.252}$ & $\textcolor{black}{0.195}$ & $\textcolor{black}{0.590}$ & $\textcolor{black}{0.261}$ & $\textcolor{black}{\mathbf{0.588}}$ \\
\bottomrule
\end{tabular}
\end{table}

\paragraph{Discussion of $U$-only.}
$U$-only is competitive with $m{\cdot}U$ on the four sampled cells, exceeding it by $0.005$--$0.021$ on three cells and matching it on Needle. We report it for completeness, but $m{\cdot}U$ is the form derived from the first-order V-term (Eq.~\eqref{eq:per_token_decomp}), and we prefer it on three grounds: (i)~it is the form predicted by the analysis, while $U$-only sets $s_j = U_j$ without any structural justification for the choice; (ii)~$m_i$ is already produced by the check-layer pass at no marginal cost, so the multiplicative gate adds cache-side fidelity grounding for free; and (iii)~the additive variant $m{+}U$ collapses to within $\pm 0.012$ F1 of Random, demonstrating that information about both factors does not by itself determine the right runtime score. The empirical near-tie between $U$-only and $m{\cdot}U$ across these four cells does not extend to all workloads: on cells where suffix attention is more diffuse, the cache-side gate is expected to provide a robustness margin that $U$-only cannot, though we leave a systematic study of this distinction to future work.
\endgroup
% Rebuttal material (oracle study, key-drift term, prompt-length sweep)
\input{appendix_rebuttal_material_camera_ready}

\clearpage % output the pending prompt-length sweep tables before the last appendix section

\section{Related Work: Full Discussion and Landscape}
\label{app:related_landscape}

This appendix expands the discussion across the literature on KV cache reuse and selective recomputation, and concludes with a full taxonomy table.

\subsection{Prefix-Aligned Reuse and Position-Independent Caching}

Prefix-aligned KV reuse~\citep{kwon2023efficient,zheng2024sglang,gim2024prompt,jin2025ragcache,liu2025lmcache,pan2026kvflow} requires identical leading token sequences and does not apply directly to RAG, where retrieved chunks vary across queries~\citep{agarwal2025cache}. A second family makes each chunk's KV cache position-independent so that chunks can be assembled into arbitrary contexts. EPIC~\citep{hu2024epic} formalizes the position-independent caching (PIC) framework, and follow-ups address related cross-context-mismatch challenges through positional reindexing~\citep{lu2025turborag,yang2026kvlink}, parallel encoding adaptation~\citep{yang2025ape}, block-wise attention masking~\citep{ma2025block}, and vision-language analogues~\citep{qin2025vlcache}. These methods reduce online prefill but rely on offline alignment or additional training rather than per-token selection at request time.

\subsection{Selective Recomputation: Token-Ranking Criteria}

Within selective recomputation, methods differ in how they rank tokens for online recomputation.

\paragraph{Deviation-based.} CacheBlend~\citep{yao2025cacheblend} ranks tokens by shallow-layer KV deviation, with the recomputation ratio set as a system parameter rather than derived from request-level harm; KVShare~\citep{yang2025kvshare} extends with a dual-stage high-deviation algorithm.

\paragraph{Boundary- and structure-based.} EPIC's LegoLink~\citep{hu2024epic} mitigates attention-sink effects at document boundaries; MEPIC~\citep{wang2025mepic} extends to memory-efficient block-level recomputation.

\paragraph{Query-, attention-, and information-flow-aware.} ProphetKV~\citep{wang2026prophetkv}, CacheClip~\citep{yang2025cacheclip}, $A^3$~\citep{zhou2026a3}, QCFuse~\citep{yan2026qcfuse}, FusionRAG~\citep{wang2026prefix}, and InfoFlow KV~\citep{teng2026infoflow} model whether reused tokens are relevant to generation through query attention, auxiliary-LLM relevance, or information-flow signals.

\paragraph{Adaptive cache management.} Cache-Craft~\citep{agarwal2025cache} studies cache management and reusable chunk-cache identification using intra- and inter-chunk attention characteristics in a RAG serving system.

KVFocus is closest to the deviation-based and query/attention-aware lines but distinct from both. It scores a token high only when it is both \emph{deviated} (large value deviation) and \emph{heavily attended} (large suffix-probe attention mass); either factor alone is insufficient. Moreover, KVFocus is not a heuristic fusion of the two signals: the multiplicative form follows from the first-order analysis (Section~\ref{sec:theory:synthesis}), and an additive combination of the same inputs performs close to random selection (Section~\ref{sec:exp_ablation}).

\subsection{Theoretical Analyses and Orthogonal Directions}
VLCache~\citep{qin2025vlcache} formally identifies cumulative non-prefix cache reuse error in vision-language generation and uses dynamic, layer-aware recomputation. \citet{cestola2026experimental} report an empirical study of reuse strategies. \citet{cancedda2024spectral} analyze attention-sink mechanisms.

A separate body of work targets KV cache compression during \emph{decoding} for memory pressure rather than prefill recomputation: StreamingLLM~\citep{xiao2024efficient}, H2O~\citep{zhang2023h2o}, SnapKV~\citep{li2024snapkv}, KVQuant~\citep{hooper2024kvquant}, SAGE-KV~\citep{wang2025llms}. These methods are orthogonal to our setting; they could in principle be combined with KVFocus to reduce both prefill cost and decode-side memory pressure. Nevertheless, eviction methods are relevant to deriving a downstream token-utility score. For instance, H2O~\citep{zhang2023h2o} retains ``heavy-hitter'' tokens, such as punctuation, that accumulate large attention, while KVzap~\citep{jegou2026kvzap} and Self-Pruned KV Attention~\citep{szilvasy2026self} learn to predict the future utility of KV entries from hidden states. Such signals could flag, already when a document is cached offline, the tokens that are likely worth recomputing, beyond the layer-1 statistics used by KVFocus; we leave this direction to future work.

\subsection{Full landscape table}

Table~\ref{tab:landscape_full} classifies the selective-recomputation methods discussed above against the first-order harm structure of Section~\ref{sec:theory:synthesis}.

\begin{table}[h]
\caption{Full landscape of selective-recomputation methods for KV cache reuse in RAG, classified against the first-order harm structure of Section~\ref{sec:theory:synthesis}.
\textbf{Mag.}: cache-side error magnitude over the reused region.
\textbf{Dep.}: downstream dependence of the suffix on the reused region.
\textbf{Per-token}: scoring resolution at the token rather than chunk level.
\checkmark{}\,$=$\,captured, $\sim$\,$=$\,partial, \textendash{}\,$=$\,absent.}
\label{tab:landscape_full}
\centering
\small
\begin{tabular}{lcccl}
\toprule
\textbf{Method} & \textbf{Mag.} & \textbf{Dep.} & \textbf{Per-token} & \textbf{Primary signal} \\
\midrule
CacheBlend~\citep{yao2025cacheblend}      & \checkmark   & \textendash & \checkmark        & Shallow-layer KV deviation \\
KVShare~\citep{yang2025kvshare}           & \checkmark   & \textendash & \checkmark        & Dual-stage KV deviation \\
EPIC~\citep{hu2024epic}                   & \textendash  & \textendash & $\sim$            & Boundary (attention sink) tokens \\
MEPIC~\citep{wang2025mepic}               & \textendash  & \textendash & \textendash       & Block-level recomputation \\
ProphetKV~\citep{wang2026prophetkv}       & \textendash  & \checkmark  & \checkmark        & Query-driven layer-wise attention \\
CacheClip~\citep{yang2025cacheclip}       & \textendash  & \checkmark  & \checkmark        & Auxiliary-LLM relevance \\
$A^3$~\citep{zhou2026a3}                   & \textendash  & \checkmark  & \checkmark        & Query-document attention \\
QCFuse~\citep{yan2026qcfuse}              & \textendash  & \checkmark  & \checkmark        & Query-centric attention summary \\
InfoFlow KV~\citep{teng2026infoflow}      & \textendash  & \checkmark  & \checkmark        & Information-flow-aware attention \\
FusionRAG~\citep{wang2026prefix}          & \textendash  & \checkmark  & \checkmark        & Query-aware fusion of pre-fused chunks \\
VLCache~\citep{qin2025vlcache}            & \checkmark   & \textendash & \checkmark        & Cumulative non-prefix reuse error \\
Cache-Craft~\citep{agarwal2025cache}      & \textendash  & $\sim$      & \textendash       & Intra/inter-chunk attention \\
\midrule
\textbf{KVFocus (ours)}                   & \textbf{\checkmark} & \textbf{\checkmark} & \textbf{\checkmark} & Value deviation $\times$ suffix attention \\
\bottomrule
\end{tabular}
\end{table}

The \textbf{Mag.} column reflects whether a method's primary signal explicitly depends on the cache-side magnitude of the K/V difference between the cached chunk and an online recomputation. CacheBlend, KVShare, and VLCache do; EPIC and the query/attention-aware family rely on relevance or structural signals that do not directly measure cache-side mismatch.
The \textbf{Dep.} column reflects whether the method's primary signal explicitly depends on downstream attention mass. The query/attention-aware family does so by design; deviation-based methods do not.
The \textbf{Per-token} column reflects the resolution at which the method scores. Most methods score per token; EPIC scores boundary tokens but treats inner tokens uniformly (\,$\sim$\,); MEPIC and Cache-Craft operate at the block or chunk level.

KVFocus is the only method whose primary signal is the explicit \emph{product} of cache-side value deviation and downstream suffix attention mass, derived from the first-order V-term decomposition of Eq.~\eqref{eq:per_token_decomp}.

\section{Limitations and Future Work}
\label{sec:discussion}% Compatibility with the unchanged Overleaf checklist.
\label{app:limitations}

\paragraph{What the analysis says.}
The first-order V-term decomposition (Eq.~\eqref{eq:per_token_decomp}) frames the selector as targeting tokens whose cache is both \emph{deviated} (large $m_i$) and \emph{heavily attended} by the suffix (large $U_i$). The ablation in Section~\ref{sec:exp_ablation} supports two structural claims of this view: rank-normalized additive composition $m{+}U$ collapses to within $\pm 0.012$ F1 of random selection, indicating that the form of combination is not interchangeable; and the multiplicative form $m{\cdot}U$ retains the strength of attention-based selection while adding a cache-side fidelity gate at essentially no marginal cost, since $m_i$ is already produced by the check-layer pass. As stated in Section~\ref{sec:theory}, the analysis serves as design motivation rather than as a tight guarantee.

\paragraph{Limitations.}
\emph{Fixed-rate selection.} The recomputation rate $r$ is a deployment hyperparameter; a per-request adaptive policy that allocates more budget to harder requests is a natural extension that we leave open.

\emph{V-term only.} The score targets only the V-term of the first-order expansion (Eq.~\eqref{eq:per_token_decomp}); key-drift correction and higher-order cross-token interactions are not addressed. Our offline analysis (Appendix~\ref{app:keyterm}) suggests that adding the key-drift term does not change F1 significantly at the evaluated rates.

\emph{Greedy decoding.} The decode-time stability result (Proposition~\ref{prop:decode}) is stated for greedy decoding and a shared decode history up to the first divergence; sampling and post-divergence dynamics require additional treatment.

\emph{Cache management and short-input workloads.} KVFocus assumes position-aligned non-prefix reuse and selects \emph{which tokens} of an already-loaded chunk to recompute; \emph{which chunks} to admit, evict, or pool is orthogonal and complementary, e.g., Cache-Craft~\citep{agarwal2025cache}. Selective recomputation is most useful when the FR$-$Naive quality gap is non-trivial; on workloads with very short inputs or already-near-FR Naive baselines (e.g., Multi-News, Section~\ref{sec:exp_efficiency}), the fixed reuse overhead can outweigh the benefit, and KVFocus does not provide a speedup (the break-even prompt length is about $0.8$K--$1.5$K tokens; Appendix~\ref{app:length_sweep}).

\begingroup
\emph{Remaining gap to full recomputation.} At low recomputation rates, KVFocus does not close the quality gap to full recomputation (Section~\ref{sec:exp_efficiency}). The oracle study (Appendix~\ref{app:oracle}) shows that part of this gap could be recovered by a better online proxy, while the rest requires a larger recomputation budget. Where latency does not force cache reuse, or where an answer error is unacceptable, full prefill remains the appropriate choice.

\emph{Granularity of the risk estimate.} The score is read at a single early layer and averaged over heads and probe positions. Since attention patterns differ across heads and layers, the layer-1, head-averaged estimate may not always reflect the attention under full recomputation; finer-grained estimation is left to future work.

\emph{Batching.} All latency measurements use batch size~1. We have not evaluated KVFocus under continuous batching, where the check-layer scoring and selective recomputation of a request are scheduled together with the prefill and decoding of other requests.
\endgroup

\paragraph{Broader impacts.}
\begingroup%
KVFocus can reduce prefill computation and user-facing latency in long-context RAG. Cache reuse can also change generated answers, as the quality comparisons show. Deployments should therefore evaluate the latency-quality trade-off on their own workloads, particularly when errors are costly. Reduced computation may lower serving energy, but we do not report an energy measurement.
\endgroup%

%% file: appendix_rebuttal_material_camera_ready.tex
% Rebuttal material for the camera-ready version: oracle study (Reviewer WwYA),
% key-drift term (Reviewer AJGS), and prompt-length sweep (Reviewer UBdB).
% The oracle and key-drift sections follow the analyses and numbers posted in the rebuttal.
\begingroup
\section{Oracle Selection Study}
\label{app:oracle}

To understand where the remaining quality gap to full recomputation (FR) lies, we construct oracle selectors that rank the reused tokens with quantities that are available only after FR has been run; these selectors therefore cannot be computed online.

\paragraph{Oracle design.}
We run FR and full reuse on the same requests and record, at the last layer, the attention and value of each reused token $j$ under both runs: $a_j$ and $a'_j$ (averaged over query heads and over the suffix-probe positions $S_p^{\mathrm{suff}}$), and $V_j$ and $V'_j$, where unprimed and primed quantities come from FR and full reuse, respectively, and $\Delta V_j = V_j - V'_j$. The reused tokens are ranked by one of four scores: (i) attention alone, $a_j$; (ii) deviation alone, $\|\Delta V_j\|_2$; (iii) their product, $a_j \|\Delta V_j\|_2$, which is the oracle counterpart of the KVFocus score; and (iv) $\|a_j V_j - a'_j V'_j\|_2$, which additionally captures the change caused by the attention weights themselves. For each score and recomputation rate, the top-ranked tokens form the recomputation set, which is replayed through the same selective-recomputation path, budget, and warmup as the deployed method. We evaluate $r \in \{1, 2, 5, 10, 15, 20, 30, 50, 80\}\%$ on Mistral-7B with 2WikiMQA, HotpotQA, and Needle-in-a-Haystack (200 samples each), using per-sample paired differences with 95\% bootstrap intervals; each rate is run independently. As a control, a mask that selects every reused token reproduces FR. The oracle in Table~\ref{tab:oracle} reports the best of the four scores at each rate.

\begin{table}[h]

\caption{ Headroom recovered (\%), i.e., $(\mathrm{F1}^{\mathrm{method}} - \mathrm{F1}^{\mathrm{Naive}})/(\mathrm{F1}^{\mathrm{FR}} - \mathrm{F1}^{\mathrm{Naive}}) \times 100$, on Mistral-7B over 2WikiMQA, HotpotQA, and Needle-in-a-Haystack. Oracle reports the best of the four oracle scores at each rate; $100\%$ corresponds to FR.}
\label{tab:oracle}
\centering
\small
\setlength{\tabcolsep}{4pt}
\begin{tabular}{l|ccccccccc}
\toprule
Method & $1\%$ & $2\%$ & $5\%$ & $10\%$ & $15\%$ & $20\%$ & $30\%$ & $50\%$ & $80\%$ \\
\midrule
Random selection & $-0.3$ & $-1.4$ & $1.5$ & $7.4$ & $9.5$ & $12.4$ & $20.8$ & $37.9$ & $77.8$ \\
KVFocus & $43.2$ & $50.0$ & $48.0$ & $61.9$ & $70.6$ & $73.8$ & $82.5$ & $93.1$ & $96.9$ \\
Oracle & $52.5$ & $56.1$ & $63.7$ & $74.1$ & $84.8$ & $91.4$ & $91.4$ & $93.2$ & $99.5$ \\
\bottomrule
\end{tabular}
\end{table}

\paragraph{Where the gap lies.}
Table~\ref{tab:oracle} reveals two gaps of different kinds. (a) \emph{Oracle versus FR.} Even with true last-layer scores, the oracle trails FR by $47.5$, $25.9$, and $8.6$ percentage points of headroom at $r = 1\%$, $10\%$, and $20\%$, respectively. A reused token that is not recomputed keeps a value computed without the current cross-chunk context; this information is absent from the cache, and no ranking can supply it. Token selection determines which tokens are repaired, not how many can be. (b) \emph{KVFocus versus the oracle.} KVFocus trails the oracle by $9.3$ points at $r = 1\%$ and by $12.2$ points at $r = 10\%$ ($12.5$ points on average over $r \le 20\%$). The difference lies in where the score is read: the oracle reads the last layer of an FR reference run, whereas KVFocus reads its score at layer~1 inside the reuse pass and adds no layer to the forward pass. A better online proxy could recover part of this gap. Closing the gap to FR therefore requires both better token selection and a larger recomputation budget: a better online selector closes part of the gap, and additional budget closes the rest.

\paragraph{Scope.}
Post-hoc combinatorial optimization over token subsets is intractable: the number of candidate subsets $\binom{N}{k}$ exceeds $10^{20}$ at our prompt lengths even at the smallest budget, and the quality of a subset is not additive over tokens, since each repaired token is recomputed while the tokens around it remain stale. Our oracles therefore bound per-token scoring given true scores, rather than the optimum over subsets. All four oracle scores select whole tokens, so the study varies the information used for selection while holding the unit of recomputation fixed.
\endgroup
\begingroup
\section{Effect of the Key-Drift Term}
\label{app:keyterm}

The first-order expansion of $\delta o_{u,C}^{(\ell)}$ (Appendix~\ref{app:proof:suffix:pertok}) contains a value term and a key term that perturbs the attention weights through $\Delta K_j^{(\ell)}$. The KVFocus score uses only the value term. We examine how much this choice affects the selected tokens and the answer quality.

\paragraph{Ranking similarity.}
We decompose the measured error exactly into the value term and the key term and compare the token rankings that the two terms induce. On Mistral-7B with 2WikiMQA (30 documents with $247{,}577$ reused tokens, layer~1), the two terms rank the tokens similarly, with a Spearman correlation of $\rho = 0.809$. Of the tokens that each term places in its top $15\%$, $73\%$ are common to both; in the top $1\%$, $63\%$ are common to both. Dropping the key term thus changes part of the selection but leaves most of it in place.

\paragraph{Effect on answer quality.}
We add the key term back into the score offline and measure the change in F1 on 2WikiMQA (Table~\ref{tab:keyterm}). None of the changes is statistically significant, and they average $+0.001$: the selection moves, but the answers do not follow it.

\begin{table}[h]

\caption{ Change in F1 on 2WikiMQA (Mistral-7B) when the key term is added back into the KVFocus score. None of the changes is statistically significant.}
\label{tab:keyterm}
\centering
\small
\begin{tabular}{l|cccccc}
\toprule
Recomputation rate $r$ & $1\%$ & $2\%$ & $5\%$ & $10\%$ & $15\%$ & $20\%$ \\
\midrule
$\Delta$F1 & $+0.012$ & $-0.014$ & $+0.014$ & $-0.026$ & $+0.020$ & $-0.001$ \\
\bottomrule
\end{tabular}
\end{table}

The key term additionally requires, for every probe and head, the projection $q_u^\top \Delta K_j$ and the centered value $V_j - o_u$ of each reused token. This adds computation at the check layer without a measurable gain in F1, so we drop the key term from the deployed score.

\endgroup

\begin{table}[!t]

\centering\small
\setlength{\tabcolsep}{4pt}
\caption{ Complete prompt-length sweep for Mistral-7B-Instruct-v0.2. FR is mean client-side TTFT in seconds. The other entries are ratios of FR mean TTFT to method mean TTFT, with values above one indicating lower latency. $n$ is the retained request count per method and length. Horizontal rules separate run families, not continuous measurements from one implementation version.}
\label{tab:length_mistral}
\begin{tabular}{rrr|rrrrr}
\toprule
Target tokens & $n$ & FR (s) & Naive & KVFocus 1\% & KVFocus 15\% & CB 15\% & EPIC 15\% \\
\midrule
256 & 9 & 0.029 & 0.654 & 0.578 & 0.556 & 0.536 & 0.583 \\
512 & 10 & 0.034 & 0.705 & 0.631 & 0.567 & 0.547 & 0.630 \\
768 & 10 & 0.048 & 0.944 & 0.765 & 0.766 & 0.779 & 0.854 \\
\midrule
1000 & 9 & 0.061 & 1.124 & 0.977 & 0.968 & 0.963 & 1.116 \\
1526 & 10 & 0.096 & 1.596 & 1.318 & 1.357 & 1.247 & 1.579 \\
2053 & 10 & 0.137 & 2.098 & 1.721 & 1.740 & 1.737 & 2.099 \\
2579 & 10 & 0.172 & 2.457 & 1.951 & 1.978 & 2.038 & 2.298 \\
3105 & 10 & 0.237 & 3.144 & 2.328 & 2.497 & 2.424 & 3.043 \\
3632 & 10 & 0.259 & 3.208 & 2.502 & 2.309 & 2.391 & 2.730 \\
4158 & 10 & 0.307 & 3.570 & 2.738 & 2.593 & 1.999 & 3.109 \\
4684 & 10 & 0.348 & 3.798 & 2.922 & 2.600 & 2.550 & 3.103 \\
5211 & 10 & 0.379 & 3.928 & 2.994 & 2.473 & 2.489 & 2.892 \\
5737 & 10 & 0.446 & 4.259 & 3.208 & 2.766 & 2.857 & 3.291 \\
6263 & 10 & 0.468 & 4.273 & 3.204 & 2.701 & 2.787 & 3.187 \\
6789 & 10 & 0.515 & 4.472 & 3.347 & 2.662 & 2.911 & 3.385 \\
7316 & 10 & 0.572 & 4.775 & 3.496 & 2.941 & 3.120 & 3.676 \\
7842 & 10 & 0.615 & 4.897 & 3.393 & 3.056 & 3.087 & 3.713 \\
8368 & 10 & 0.678 & 5.199 & 3.666 & 3.239 & 3.347 & 3.981 \\
8895 & 10 & 0.717 & 5.248 & 3.816 & 2.839 & 2.793 & 3.395 \\
9421 & 10 & 0.756 & 5.267 & 3.858 & 2.891 & 2.959 & 3.486 \\
9947 & 10 & 0.817 & 5.548 & 3.945 & 3.047 & 3.124 & 3.699 \\
10474 & 10 & 0.858 & 5.639 & 3.815 & 2.975 & 3.070 & 3.505 \\
11000 & 10 & 0.899 & 5.734 & 3.911 & 2.971 & 3.116 & 3.687 \\
\midrule
13000 & 9 & 1.088 & 6.108 & 4.393 & 2.815 & 2.912 & 3.430 \\
16000 & 10 & 1.423 & 6.772 & 4.747 & 3.143 & 3.136 & 3.834 \\
\bottomrule
\end{tabular}
\end{table}

\begin{table}[!t]

\centering\small
\setlength{\tabcolsep}{4pt}
\caption{ Complete prompt-length sweep for Qwen2.5-7B-Instruct. FR is mean client-side TTFT in seconds. The other entries are ratios of FR mean TTFT to method mean TTFT, with values above one indicating lower latency. $n$ is the retained request count per method and length. Horizontal rules separate run families, not continuous measurements from one implementation version.}
\label{tab:length_qwen25}
\begin{tabular}{rrr|rrrrr}
\toprule
Target tokens & $n$ & FR (s) & Naive & KVFocus 1\% & KVFocus 15\% & CB 15\% & EPIC 15\% \\
\midrule
256 & 9 & 0.028 & 0.640 & 0.561 & 0.585 & 0.587 & 0.614 \\
512 & 10 & 0.029 & 0.624 & 0.404 & 0.507 & 0.543 & 0.583 \\
768 & 10 & 0.046 & 0.936 & 0.770 & 0.723 & 0.767 & 0.824 \\
\midrule
1000 & 9 & 0.062 & 1.225 & 1.083 & 1.047 & 1.033 & 1.122 \\
1526 & 10 & 0.097 & 1.742 & 1.491 & 1.397 & 1.482 & 1.650 \\
2053 & 10 & 0.134 & 2.233 & 1.874 & 1.816 & 1.812 & 2.096 \\
2579 & 10 & 0.160 & 2.488 & 2.035 & 1.971 & 2.016 & 2.325 \\
3105 & 10 & 0.189 & 2.748 & 2.247 & 2.166 & 2.231 & 2.571 \\
3632 & 10 & 0.241 & 3.304 & 2.646 & 2.434 & 2.559 & 2.951 \\
4158 & 10 & 0.281 & 3.623 & 2.839 & 2.572 & 2.681 & 3.253 \\
4684 & 10 & 0.326 & 3.994 & 3.054 & 2.879 & 2.928 & 3.500 \\
5211 & 10 & 0.375 & 4.369 & 3.193 & 2.415 & 2.687 & 3.217 \\
5737 & 10 & 0.399 & 4.248 & 3.217 & 2.712 & 2.659 & 3.301 \\
6263 & 10 & 0.429 & 4.372 & 3.261 & 2.786 & 2.595 & 3.277 \\
6789 & 10 & 0.488 & 4.749 & 3.562 & 2.983 & 3.189 & 3.670 \\
7316 & 10 & 0.519 & 4.875 & 3.623 & 3.003 & 3.053 & 3.628 \\
7842 & 10 & 0.558 & 5.043 & 3.535 & 2.595 & 3.073 & 3.658 \\
8368 & 10 & 0.617 & 5.355 & 3.818 & 3.141 & 3.302 & 3.936 \\
8895 & 10 & 0.661 & 5.506 & 3.855 & 3.024 & 3.360 & 4.025 \\
9421 & 10 & 0.703 & 5.656 & 3.935 & 3.275 & 3.377 & 4.077 \\
9947 & 10 & 0.758 & 5.825 & 4.173 & 3.442 & 3.548 & 4.298 \\
10474 & 10 & 0.794 & 5.979 & 4.185 & 3.028 & 2.998 & 3.634 \\
11000 & 10 & 0.831 & 6.065 & 3.561 & 3.114 & 3.182 & 3.792 \\
\midrule
13000 & 9 & 0.986 & 6.094 & 4.325 & 3.143 & 3.047 & 3.825 \\
16000 & 10 & 1.280 & 6.702 & 4.099 & 3.136 & 3.111 & 3.801 \\
\bottomrule
\end{tabular}
\end{table}

\begin{table}[!t]

\centering\small
\setlength{\tabcolsep}{4pt}
\caption{ Complete prompt-length sweep for Qwen3-8B. FR is mean client-side TTFT in seconds. The other entries are ratios of FR mean TTFT to method mean TTFT, with values above one indicating lower latency. $n$ is the retained request count per method and length. Horizontal rules separate run families, not continuous measurements from one implementation version.}
\label{tab:length_qwen3}
\begin{tabular}{rrr|rrrrr}
\toprule
Target tokens & $n$ & FR (s) & Naive & KVFocus 1\% & KVFocus 15\% & CB 15\% & EPIC 15\% \\
\midrule
256 & 9 & 0.032 & 0.586 & 0.564 & 0.539 & 0.498 & 0.563 \\
512 & 10 & 0.032 & 0.551 & 0.470 & 0.466 & 0.450 & 0.518 \\
768 & 10 & 0.049 & 0.772 & 0.688 & 0.682 & 0.611 & 0.763 \\
\midrule
1000 & 9 & 0.068 & 1.071 & 1.001 & 0.993 & 0.892 & 1.069 \\
1526 & 10 & 0.100 & 1.425 & 1.313 & 1.316 & 1.282 & 1.417 \\
2053 & 10 & 0.136 & 1.795 & 1.606 & 1.644 & 1.628 & 1.784 \\
2579 & 10 & 0.186 & 2.269 & 2.016 & 2.039 & 2.052 & 2.240 \\
3105 & 10 & 0.225 & 2.564 & 2.227 & 2.285 & 2.272 & 2.541 \\
3632 & 10 & 0.258 & 2.737 & 2.325 & 2.350 & 2.389 & 2.672 \\
4158 & 10 & 0.317 & 3.117 & 2.689 & 2.710 & 2.748 & 3.136 \\
4684 & 10 & 0.351 & 3.225 & 2.777 & 2.511 & 2.555 & 2.907 \\
5211 & 10 & 0.403 & 3.522 & 3.012 & 2.593 & 2.640 & 2.974 \\
5737 & 10 & 0.453 & 3.676 & 3.084 & 2.774 & 2.831 & 3.251 \\
6263 & 10 & 0.484 & 3.801 & 3.125 & 2.751 & 2.807 & 3.210 \\
6789 & 10 & 0.524 & 3.891 & 3.094 & 2.797 & 2.878 & 3.291 \\
7316 & 10 & 0.587 & 4.137 & 3.200 & 2.917 & 2.887 & 3.420 \\
7842 & 10 & 0.634 & 4.340 & 3.533 & 3.050 & 3.092 & 3.622 \\
8368 & 10 & 0.665 & 4.350 & 3.536 & 3.095 & 3.088 & 3.715 \\
8895 & 10 & 0.739 & 4.638 & 3.778 & 2.819 & 2.862 & 3.282 \\
9421 & 10 & 0.783 & 4.772 & 3.848 & 2.894 & 2.936 & 3.386 \\
9947 & 10 & 0.858 & 5.012 & 4.027 & 3.057 & 2.970 & 3.560 \\
10474 & 10 & 0.895 & 5.069 & 4.058 & 2.995 & 3.035 & 3.521 \\
11000 & 10 & 0.938 & 5.064 & 4.077 & 2.879 & 2.900 & 3.601 \\
\midrule
13000 & 9 & 1.163 & 5.749 & 3.624 & 2.614 & 2.668 & 3.338 \\
16000 & 10 & 1.516 & 6.456 & 4.539 & 3.143 & 3.278 & 3.731 \\
\bottomrule
\end{tabular}
\end{table}

\begingroup%
\section{Prompt-Length Sweep}
\label{app:length_sweep}

We report the complete prompt-length sweep on the Needle-in-a-Haystack task over 25 target context lengths from 256 to 16{,}000 tokens for all three models. The sweep uses 256-token chunks and compares FR, naive reuse, KVFocus at $1\%$ and $15\%$, CacheBlend at $15\%$, and EPIC at $15\%$. Tables~\ref{tab:length_mistral}--\ref{tab:length_qwen3} give the reported FR mean TTFT and each reuse method's speedup.

\paragraph{Timing protocol.}
For each method and run, we discard only the first request, then compute speedup as the ratio of the FR mean TTFT to the method's mean TTFT at the same target length. This is a ratio of means, not a mean of per-request speedups. The first length in each run family therefore retains nine requests, and the other lengths retain ten.

The short (256--768), middle (1{,}000--11{,}000), and long (13{,}000--16{,}000) ranges come from three run families collected with different implementation versions. We compare methods within the corresponding family. These measurements are not pooled with the main deployment timings, and changes across family boundaries should not be attributed solely to context length.

\paragraph{Short-input overhead.}
At 256--768 target tokens, all tested reuse methods, including naive reuse, have speedup below one. This is consistent with fixed reuse overhead dominating the prefill computation saved on short inputs. For KVFocus at $1\%$, the recorded speedup first reaches or exceeds one between 768 and 1{,}000 tokens for the Qwen models, and between 1{,}000 and 1{,}526 for Mistral. Qwen3 is approximately at parity at 1{,}000 tokens. These break-even intervals are between tested points. Since naive reuse, which recomputes nothing, is also slower than FR at these short lengths, this overhead stems from the reuse serving path rather than from the check-layer scoring of KVFocus. Beyond the break-even point, the speedup keeps increasing with the prompt length; RAG prompts that concatenate multiple retrieved documents typically lie far above it. The sweep measures latency only and provides no quality comparison.

\endgroup%